\documentclass[10pt,journal,compsoc]{IEEEtran}
\ifCLASSOPTIONcompsoc
  \usepackage[nocompress]{cite}
\else
  \usepackage{cite}
\fi
\usepackage{amsfonts}       % blackboard math symbols
\usepackage{multirow}
\usepackage[captionskip=2pt]{floatrow}
\usepackage[skip=2pt,margin=2pt,font=footnotesize]{caption}
\newfloatcommand{capbtabbox}{table}[][\FBwidth]  
\usepackage{blindtext}  
\usepackage[numbers,sort, compress]{natbib}

\usepackage{graphicx}

\usepackage{epsfig}
\usepackage{amsmath}
\usepackage{amssymb}
\usepackage{url}
\usepackage[font=footnotesize,skip=0pt]{caption}

\begin{document}
%\title{A Representation of Image Semantics for Object to Scene Transfer}
\title{Semantic Fisher Scores for Task Transfer: \\
  Using Objects to Classify Scenes}

\author{
  Mandar~Dixit,~\IEEEmembership{Student Member,~IEEE,}
  Yunsheng~Li,~\IEEEmembership{Student Member,~IEEE,}
  Nuno~Vasconcelos,~\IEEEmembership{Fellow,~IEEE,}% <-this % stops a space
\IEEEcompsocitemizethanks{\IEEEcompsocthanksitem M. Dixit is with the Microsoft, Redmond, WA, 98052.\protect\\
E-mail: madixit@microsoft.com
\IEEEcompsocthanksitem Y. Li and N. Vasconcelos are with Department of Electrical and Computer Engineering, University of California at San Diego, La Jolla, CA 92093.}% <-this % stops an unwanted space
\thanks{Manuscript received Oct -, 2018; revised Jan -, 2019.}}


\markboth{Journal of \LaTeX\ Class Files,~Vol.~14, No.~8, August~2015}%
{Shell \MakeLowercase{\textit{et al.}}: Bare Demo of IEEEtran.cls for Computer Society Journals}



\IEEEtitleabstractindextext{%
\begin{abstract}
The tranfer of a neural network (CNN) trained to recognize objects
to the task of scene classification is considered. A Bag-of-Semantics (BoS) 
representation is first induced, by feeding scene image patches to the object
CNN, and representing the scene image by the ensuing bag of posterior class
probability vectors (semantic posteriors). The encoding of the BoS with 
a Fisher vector (FV) is then studied. A link is 
established between the FV of any probabilistic model and the $Q$-function 
of the expectation-maximization (EM) algorithm used to estimate its 
parameters by maximum likelihood. This enables 1) imediate derivation of 
FVs for any model for which an EM algorithm exists, and 2) 
leveraging efficient implementations from the EM literature for the 
computation of FVs. It is then shown that standard FVs, such as those derived 
from Gaussian or even Dirichelet mixtures, are unsuccessful for the
transfer of semantic posteriors, due to the highly non-linear nature of the 
probability simplex. The analysis of these FVs shows 
that significant benefits can ensue by 1) designing FVs 
in the natural parameter space of the multinomial distribution, and 2) adopting 
sophisticated probabilistic models of semantic feature covariance. 
The combination of these two insights leads to the encoding of the BoS in 
the natural parameter space of the multinomial, using a 
vector of Fisher scores derived from a mixture of factor analyzers (MFA). 
A network implementation of the MFA Fisher Score (MFA-FS), denoted as the 
MFAFSNet, is finally proposed to enable end-to-end training. 
Experiments with various object CNNs and datasets show 
that the approach has state-of-the-art transfer performance. Somewhat 
surprisingly, the scene classification results are superior to those of a 
CNN explicitly trained for scene classification, using a large scene 
dataset (Places). This suggests that holistic analysis
is insufficient for scene classification. The modeling of local object 
semantics appears to be at least equally important. The two
approaches are also shown to be strongly complementary, leading
to very large scene classification gains when combined, and 
outperforming all previous scene classification
approaches by a sizeable margin.
\end{abstract}

}

\maketitle

\IEEEdisplaynontitleabstractindextext

% \fi
\IEEEpeerreviewmaketitle

\IEEEraisesectionheading{\section{Introduction}\label{sec:introduction}}

\IEEEPARstart{C}{o}nvolutional neural networks~(CNNs)
have achieved remarkable
performance on vision problems such as image
classification~\cite{krizhevsky12,simonyan14,szegedy14} or object detection and
localization~\cite{girshick14,ren15,zhou14}. Beyond 
impressive results, they have an
unmatched resilience to dataset
bias~\cite{torralba11}. It is now well known that a
network trained to solve a task on a certain dataset (e.g. object
recognition on ImageNet~\cite{dset:ImNet}) can be easily 
fine-tuned to a related
problem on another dataset (e.g. object detection on
MS-COCO). Less studied is robustness to task bias, i.e.
generalization across tasks. 
% Given the large number of possible
% vision tasks, it is impossible to train a CNN from scratch for each. In fact,
% it is likely not even feasible to collect the large number of images needed
% for such training. This motivates the study of knowledge transfer across tasks
In this work, we consider an important class of such problems, where a
classifier trained on a set of semantics is transferred to a second set
of semantics, which are loose combinations of the original ones.
We consider the particular case where original semantics are
object classes and target semantics are scene classes that somehow depend
on those objects.  %Examples of this problem 
%include the transfer of object classifiers to scene 
%classification~\cite{donahue14,gong14,liu14,cimpoi15} or 
%image captioning~\cite{vinyals15}. 

Task transfer has been a topic of significant interest in computer vision. 
Prominent examples of cross-task transfer 
include object detectors learned from object recognition 
models~\cite{girshick14,he_sppnet14}, object recognizers based on 
attribute detectors~\cite{lampert09,akata16} and complex activity 
recognition methods based on attribute detection~\cite{liu_attr11,li17} or 
object recognition~\cite{jain15,jain_o2a15}. Our particular interest in 
object to scene transfer stems from the complex relation between the two 
domains. A scene can be described as a collection of multiple objects 
occurring in an unpredictable layout. Localizing the scene semantics
is already a difficult task. This is compounded by the difficulty of
mapping localized semantics into a holistic scene representation. 
The problem of knowledge transfer from object to 
scene recognizers is therefore very challenging.    

One might argue that instead of using transfer, a scene classifier CNN can 
be trained directly from a large dataset of scene images. This 
approach has two major limitations.
%While scene classification CNNs can be trained with large datasets of scene 
%images, this has two
%limitations. 
First, it does not leverage all the work already devoted to object
recognition in the literature. Both datasets and models have to be
designed from scratch, which is time consuming. Second, the ``directly learned''
CNN does not necessarily model relations between holistic scene descriptions 
and scene objects. This can degrade classification performance.
We consider instead the prediction of holistic scene tags from the scores 
produced by an object CNN classifier. Since it leverages available object 
recognition CNNs this type of transfer is more efficient in terms of data 
collection and training. We show that it can also produce better scene 
classification results. This is because a scene  classifier can leverage 
the recognition of certain types of rocks, tree stumps, or lizard species 
to distinguish between ``Arizona Desert'' and ``Joshua Tree National Park''. 
A holistically trained CNN can have difficulty honing in on these objects 
as discriminators between the two classes.
%This has, for example, been documented in the scene classification literature, where the performance of the best “directly learned” CNNs~\cite{zhou14}, can be substantially improved by fusion with object recognition CNNs~\cite{dixit15}.

\begin{figure*}[t]\RawFloats
\centering
  \includegraphics[width=.9\linewidth]{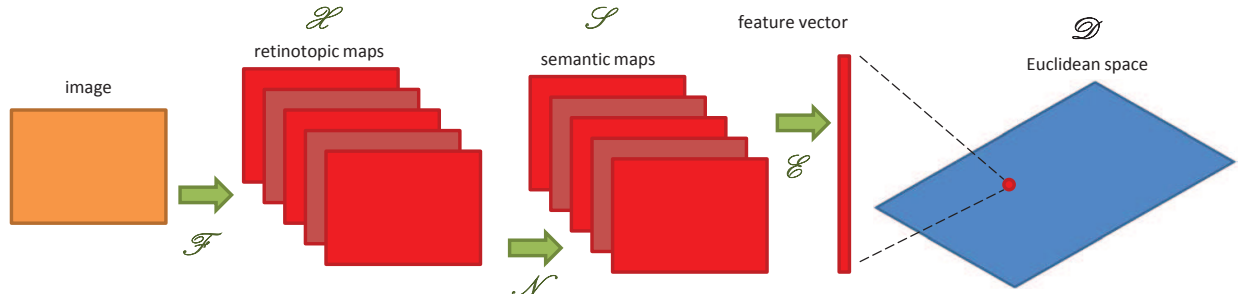}
  \caption{The bag of semantics (BoS) classifier consists of a retinotopic 
    feature mapping $\cal F$ followed by a semantic mapping $\cal N$. A 
    non-linear embedding $\cal E$ of the semantic maps is then used to 
    generate a feature vector on a Euclidean space $D$.}
%    The space $\cal X$ of retinotopic
%    features is first mapped into a retinotopic semantic space $\cal S$,
%    using a classifier of image patches. A non-linear embedding $\cal E$ is
%    then used to map this representation into a feature vector on an
%    Euclidean space $\cal D$.}
    %\vspace{-0.1in}
  \label{fig:BoS}
\end{figure*}

The proposed object-to-scene transfer is based on the {\it bag of semantics\/}
(BoS) representation. It derives a scene representation by scoring
a set of image patches with a pre-trained object classifier. The
probabilities of different objects are the {\it scene semantics\/}
and the set of probability vectors the BoS. A holistic scene classifier is
then applied to the BoS,  to transfer knowledge from objects to scenes.
Several authors have argued for semantic image representations in
vision~\cite{vogel07,rasiwasia07,su12,kwitt12,lampert13,bergamo14,
feifei-obank14}. They have been used to
describe objects by their attributes~\cite{lampert13}, represent scenes
as collections of objects~\cite{feifei-obank14, kordumova2016pooling} 
and capture contextual
relations between classes~\cite{rasiwasia11}.
For tasks such as hashing or large scale retrieval,
a global semantic descriptor is usually preferred~\cite{torresani10,bergamo11}. Works on zero-shot object based scene representation~\cite{kordumova2016pooling} also use a global semantic image descriptor, mainly because the object-to-scene transfer functions used in these problems require dimensions of the descriptor to be interpretable as object scores. Proposals for scene classification, on the other hand, tend to rely on the BoS~\cite{su12,kwitt12,feifei-obank14}. However,
while the BoS outperforms low-level features in low
dimensions~\cite{kwitt12}, it has been less effective for
high dimensional descriptors such as the Fisher vector (FV)~\cite{perronnin10}.
This is because region semantics can be noisy, and it
is hard to map a bag of probability vectors into a high dimensional scene
representation, such as a FV~\cite{perronnin10}. 

In this work, we leverage the high accuracy of ImageNet trained 
CNNs~\cite{krizhevsky12,simonyan14} to overcome
the first problem. We obtain a BoS by using these networks to
extract semantic descriptors (object class posterior probability
vectors) from local image patches. We then extend
the FV to this BoS. This {\it semantic Fisher vector\/} amounts to a large 
set of non-linear pooling operators that act on high-dimensional probability 
vectors.
We show that, unlike low-level features, this extension cannot be 
implemented by the classical Gaussian mixture model FV (GMM-FV). 
We simplify the derivation of FVs for other models, by 
linking the FV of any probabilistic model to the 
$Q$-function of the expectation-maximization (EM) algorithm used to estimate 
its parameters. It is shown that
the FV can be trivially computed as a combination of the E and M steps
of EM. This link also enables the leveraging of efficient EM 
implementations to compute FVs. It is, however, shown that even a more
natural distribution for probability vectors, the Dirichlet mixture model
(DMM), fails to generate an effective FV for the BoS. 

We hypothesize that this
is due to the non-Euclidean nature of the probability simplex,
which makes the modeling of probability distributions quite complex.
Since the FV is always defined with respect to a reference probability
distribution, this hurts classification performance.
For the GMM-FV, the problem is that the assumption of a locally 
Euclidean geometry is not well suited for image semantics, 
which are defined on the simplex.
For the DMM-FV, the problem is a lack of explicit modeling of 
second order statistics of these semantics. Nevertheless, 
an analysis of the DMM-FV
reveals a non-linear log embedding that maps
a multinomial distribution to its natural parameter space,
where the Euclidean assumption is effective. This suggests
 using a GMM model on the 
natural parameter space of the semantic multinomial (SMN), leading to 
the logSMN-FV.
In fact, because the multinomial has various natural space 
parametrization,  we seek the one best suited for 
CNN semantics. This turns out to be the inverse of the softmax implemented at 
the network output. Since the CNN is optimized for these 
semantics, this parameterization has the benefits of end-to-end training. 
It is shown that a GMM-FV of the pre-softmax CNN outputs significantly 
outperforms the GMM-FV and the DMM-FV.

% of the probability vectors into a space
% amenable to pooling, a centroid function $\xi(.)$ that
% localizes the pooling operators, and a scaling function $\gamma(.)$ that 
% normalizes the results of this embedding. 
% An empirical study shows that significant gains can be obtained by
% mixing the logarithmic embedding of the DMM-FV with the variance scaling of
% the GMM-FV. This is explained by the highly non-Euclidean nature of
% the probability simplex. On one hand, the GMM-FV assumption of a locally 
% Euclidean geometry is not well suited for image semantics, which are 
% parameters of a multinomial distribution. The log embedding 
% of the DMM-FV maps the multinomial to its natural parameter
% space, where the Euclidean assumption is effective. On the
% other hand, the combination of this mapping with GMM modeling enables
% the very effective Gaussian scaling, which is not accessible to the DMM.

While these results show an advantage for modeling second order statistics, 
the use of a GMM of diagonal covariances limits the ability of the GMM-FV 
to approximate the non-linear manifold of CNN natural parameter features. 
For this,
we resort to a richer generative model, the mixture
of factor analyzers~(MFA)~\cite{ghahramani97,verbeek06}, which locally
approximates the natural-space BoS manifold by a set of low-dimensional
linear subspaces, derived from covariance
information. We derive the MFA Fisher score (MFA-FS) and corresponding MFA-FV
and show that the covariance statistics captured by these descriptors are
highly discriminant for CNN semantics, significantly outperforming the
GMM-FV. To allow end-to-end training, the MFA-FS is finally
implemented as a neural network layer. The resulting MFAFSNet is an object
to scene transfer network that can be fine-tuned for scene classification
by backpropagation. This further improves scene classification 
performance.

Experiments on the SUN~\cite{dset:MITSUN} and 
MIT Indoor~\cite{dset:MITIndoor} datasets show that the MFA 
representations (MFA-FS and MFAFSNet) outperform
scene classifiers based on lower level CNN 
features~\cite{gong14,cimpoi15,liu14,liu16}, alternative approaches 
for second order pooling of CNN semantics~\cite{lin15,gao16}, and even 
CNNs learned directly from scene 
datasets~\cite{zhou14,khan2017scene,guo2017locally}. This is surprising, since the MFA representations
perform task transfer, applying object recognition 
CNNs to scene classification, and require little scene training data. 
This is unlike direct CNN 
training, which requires a much larger scene dataset, such as 
Places~\cite{zhou14}. Furthermore, the two representations are
complementary: combination of the MFA-FS and the scene CNN
significantly outperforms the methods in isolation. The combined
classifier has state-of-the-art scene classification performance,
achieving sizable improvements over all previous approaches.

 

\section{Bag of Semantics Classification}

We start by reviewing the foundations of BoS 
classification.

\begin{figure}[t]
\centering
  \includegraphics[scale=0.4]{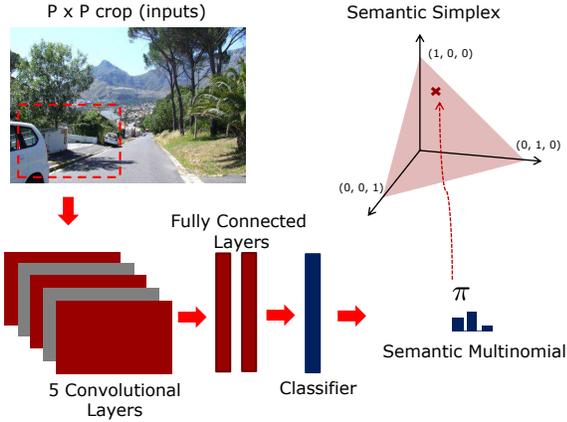}
\caption{CNN based semantic image representation. Each
  image patch is mapped into an SMN $\pi$.
}
\label{fig:simplex}
\end{figure}

\subsection{Prior work}

Figure~\ref{fig:BoS} presents the architecture of the BoS
classifier. Given an image $I(l)$, where $l$ denotes spatial 
location, it defines an initial mapping ${\cal F}$ into a set of 
{\it retinotopic\/} feature maps $f_k(l)$. These preserve spatial 
topology of the image and encode local visual information. 
%They can be seen as a (potentially non-linear) local convolution of the 
%$I(l)$ with filters that detect edges, textures or even object parts.
%In the literature, the maps $\cal F$ 
They have been implemented with 
handcrafted descriptors such as SIFT, HoG or the convolutional layers of a CNN. 
The next stage is a second retinotopic mapping $\cal N$ into the space of 
classifier outputs $\cal S$. Classifiers that define this mapping are 
pre-trained on an auxiliary set of {\it semantic concepts\/}, e.g. 
objects~\cite{feifei-obank10,feifei-obank14} or themes~\cite{rasiwasia08,rasiwasia11,kwitt12}, that occur locally within images. At each location $l$, they 
map the descriptors extracted at $l$ into 
a {\it semantic vector\/} in $\cal S$, whose entries are 
probabilities of occurrence of the individual semantic concepts. The image 
is thus, transformed into a collection or a ``bag'' of semantics. However, 
due to their retinotopic nature, a BoS is sensitive to 
variations in scene layout. If an object changes position in the field of 
view, the semantic feature map will change completely. To guarantee
{\it invariance,\/} the BoS is embedded into a fixed length 
non-retinotopic representation, using a non-linear mapping ${\cal E}$ into 
a high dimensional feature space ${\cal D}$. The space ${\cal D}$ must have 
a Euclidean structure that supports classification with linear decision 
boundaries. 

Prior BoS scene classifiers~\cite{rasiwasia08,rasiwasia11,kwitt12,
feifei-obank10,feifei-obank14} had 
limited success, for two reasons. First, scene semantics 
are non-trivial to {\it localize\/}. Scenes are collections of objects 
and stuff~\cite{adelson01} in diverse layouts. Detecting 
these entities can be challenging. Object detectors
based on handcrafted features, such as SIFT or HoG, lacked 
discriminative power, producing mappings $\cal N$ riddled
with {\it semantic noise\/}~\cite{rasiwasia11}. Second, it can be 
difficult to design an invariant scene descriptor (embedding $\cal E$). 
The classical pooling of the bag of descriptors into a vector of 
statistics works well for low and mid-level 
features~\cite{csurka04,lazebnik06,yang09} but is far less effective for the 
BoS. Semantic features are class probabilities that inhabit a very 
non-Euclidean simplex. Commonly used statistics, such as average or 
max pooling~\cite{kwitt12,feifei-obank14}, 
do not perform well in this space. Our experiments show that even 
sophisticated non-linear embeddings, such as FVs~\cite{perronnin10},
can perform poorly.  

The introduction of deep CNNs~\cite{krizhevsky12,simonyan14,szegedy14}
has all but 
solved the problem of noisy semantics. These models learn highly 
discriminative and non-linear image mappings $\cal F$ that are far 
superior to handcrafted features. Their top layers have been shown 
selective of semantics such as faces and object parts~\cite{zeiler14}.
As discussed in the following section, scoring the local regions of a 
scene with an object recognition CNN produces a robust BoS. It remains to 
%To implement a robust semantic mapping $\cal N$, therefore, an imageNet CNN can be used directly as a scene patch classifier.
design the embedding $\cal E$.
This is discussed in the remainder of the paper.

\subsection{CNN semantics}

%We start with a brief review of a BoS image representation
%and then propose suitable embeddings for them.

Given a vocabulary ${\cal V} = \{v_1, \ldots, v_S\}$ of $S$
{\it semantic concepts\/}, an image $I$ can be described as a bag of
instances from these concepts, localized within image patches/regions.
Defining an $S$-dimensional binary indicator vector
$s_i$, such that $s_{ir} = 1$ and $s_{ik} = 0$, $k \neq r$,
when the $i^{th}$ image patch $b_i$ depicts the semantic class $r$,
the image can be represented as $I = \{s_1, s_2, \ldots, s_n\}$,
where $n$ is the total number of patches.
Assuming that $s_i$ is sampled from a multinomial distribution of
parameter $\pi_i,$ the log-likelihood of $I$ is
\begin{equation}
\label{eq:smn-lkd}
{\cal L\/} = \log \prod_{i=1}^n \prod_{r=1}^S {\pi_{ir}}^{s_{ir}}
= \sum_{i=1}^N\sum_{r=1}^S {s_{ir}}\log{\pi_{ir}}.
\end{equation}
Since the semantic labels $s_i$ for image regions are unknown, it is common 
to rely instead on the expected log-likelihood
\begin{equation}
\label{eq:smn-exp-lkd}
E[{\cal L\/}] =  \sum_{i=1}^n\sum_{r=1}^S E[s_{ir}]\log{\pi_{ir}}
\end{equation}
where $E[s_{ir}] = P(r|b_i)= \pi_{ir}$ are the scene semantics for 
patch $i$, and \eqref{eq:smn-exp-lkd} depends
only on the multinomial parameters $\pi_i$.
%The expression is maximized by the choice of $\pi_i = E[s_{i}]$,
%i.e. the posterior probability vector inferred from the $i^{th}$ patch.
This is denoted the semantic multinomial~(SMN) in~\cite{rasiwasia07}.
SMNs are computed by applying a classifier,
trained on the semantics of $\cal V$, to the image patches $b_i$, and using
the resulting posterior class probabilities as $\pi_i$.
This is illustrated in Figure~\ref{fig:simplex}, for a CNN classifier.
Each patch is mapped into the probability simplex, 
denoted the semantic space $\cal S$ in Figure~\ref{fig:BoS}. The
image is finally represented by the SMN collection
$I = \{\pi_1, \ldots, \pi_n\}$. This is the BoS.

\begin{figure}[t]
  \centering
  \begin{tabular}{cc}
    \includegraphics[height=2.75cm,width=.39\linewidth]{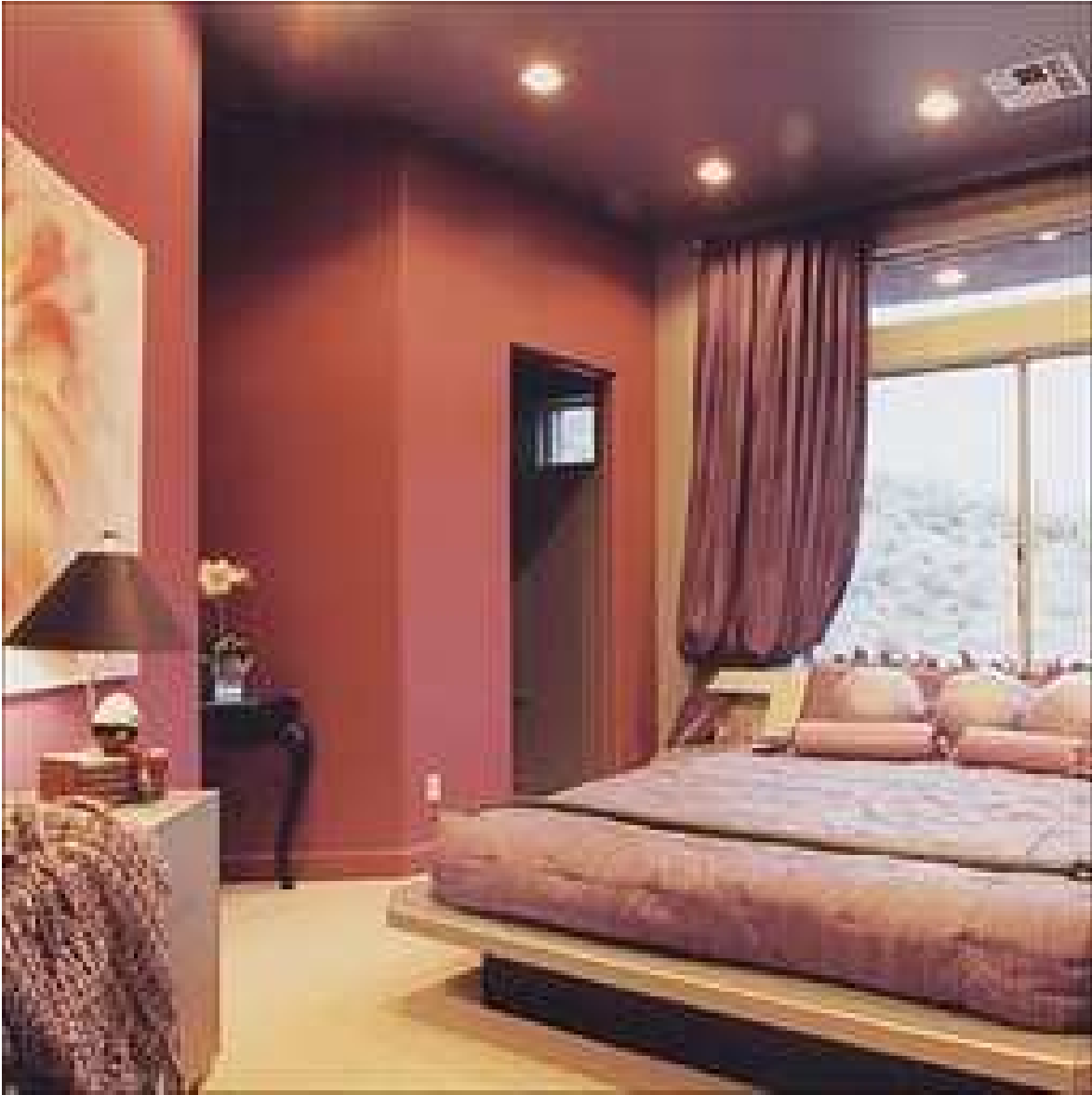} &
    \includegraphics[width=.39\linewidth]{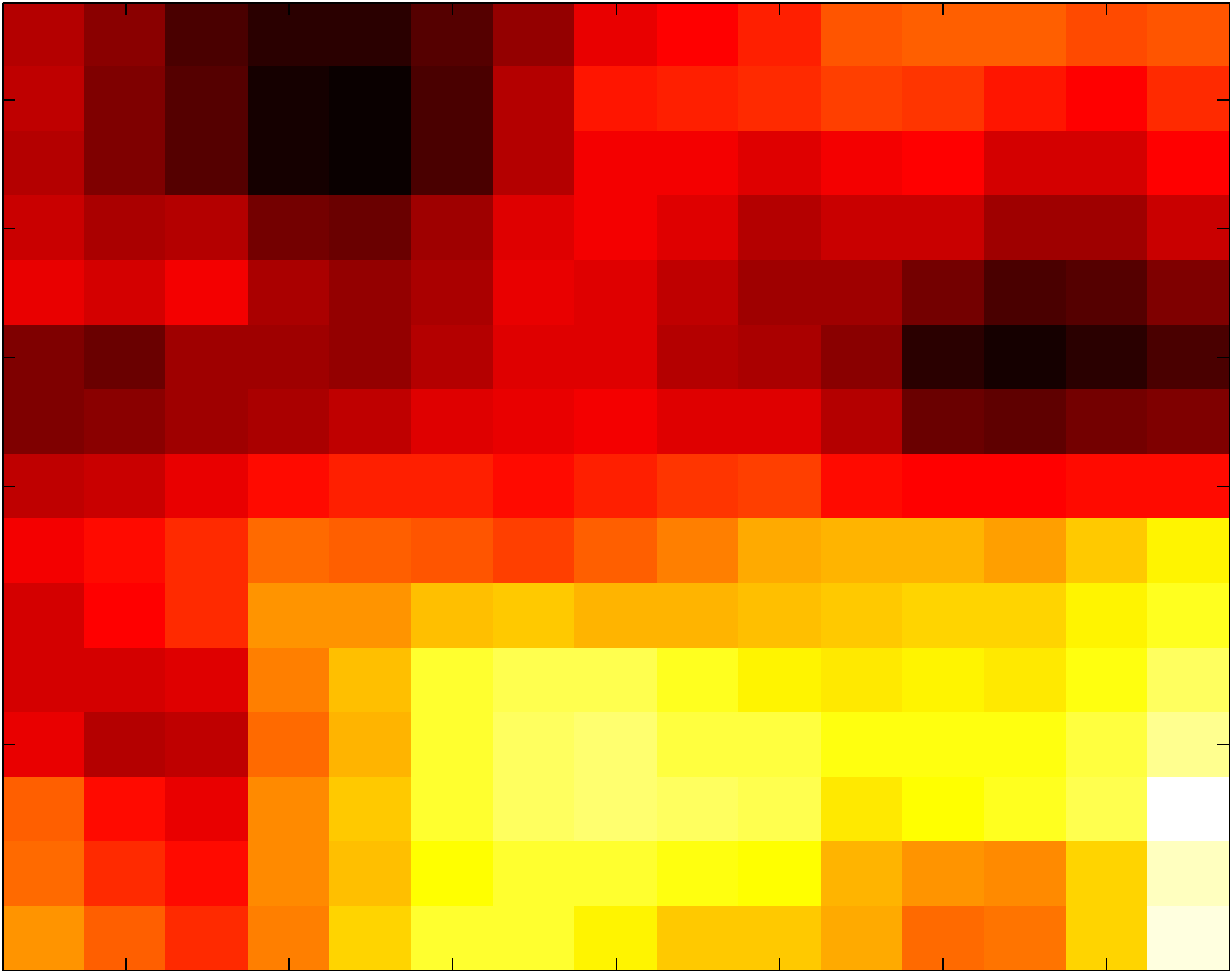} \\
    \footnotesize{a) bedroom scene} &     \footnotesize{b) ``day bed''}\\
    \includegraphics[width=.39\linewidth]{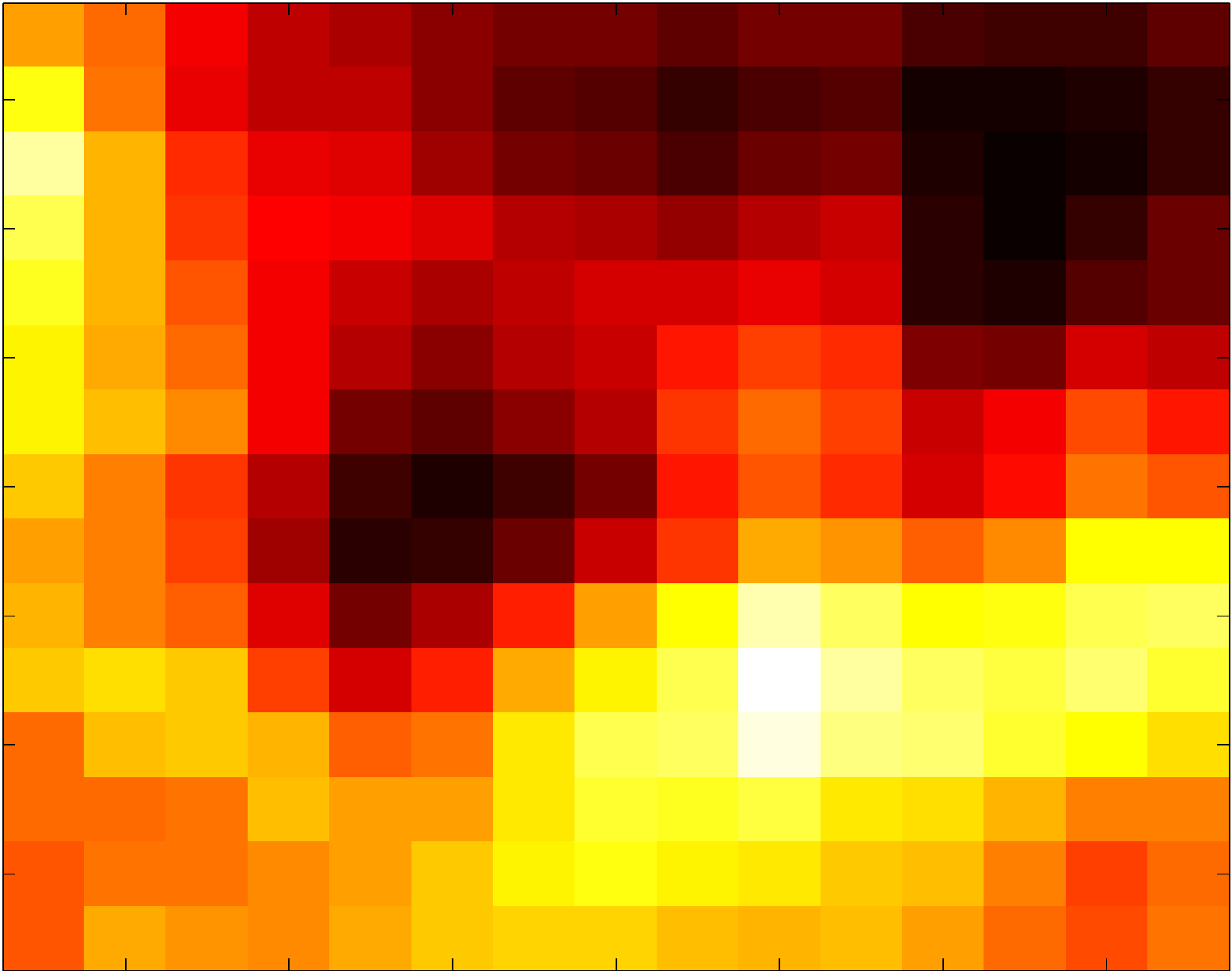} &
    \includegraphics[width=.39\linewidth]{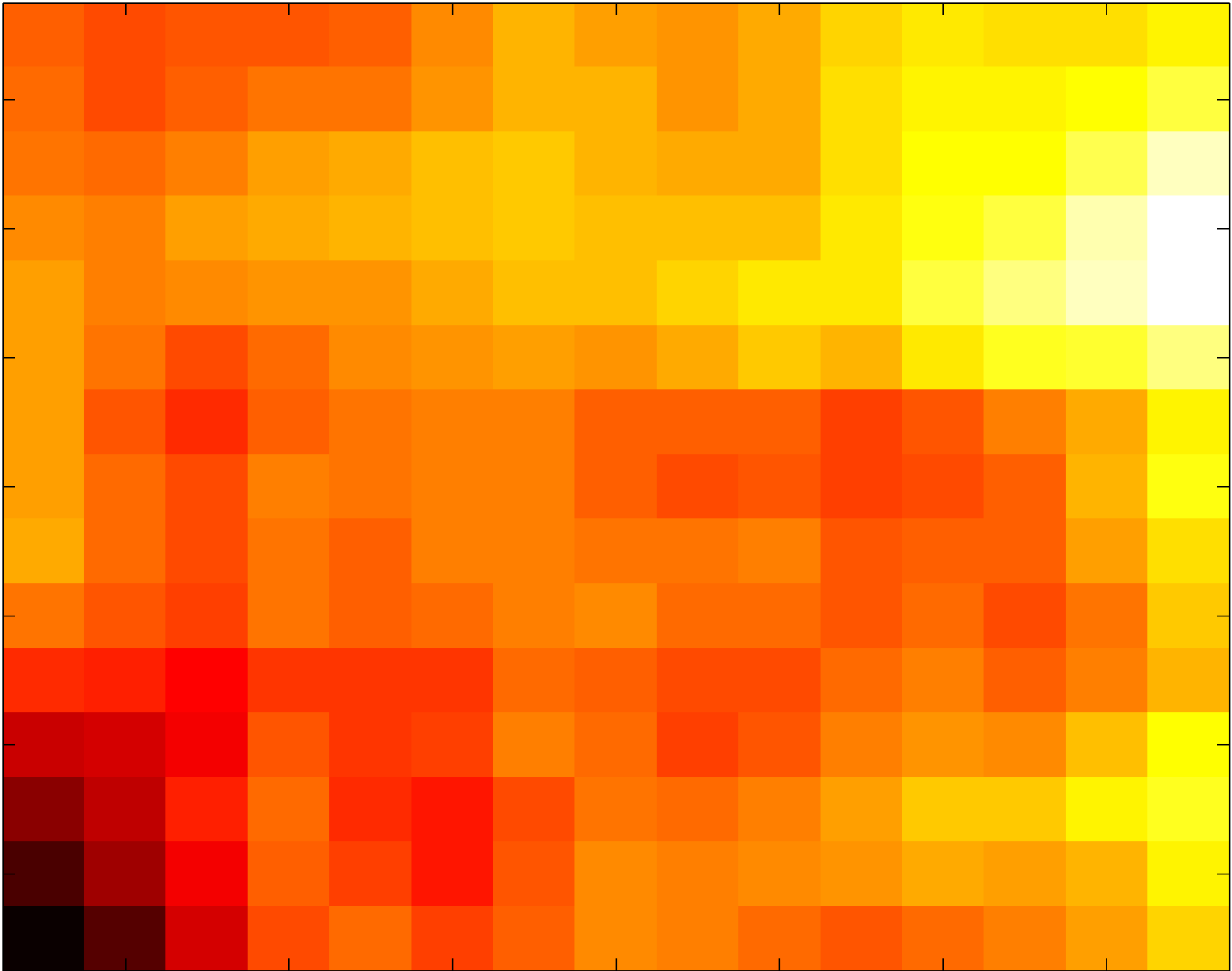} \\
    \footnotesize{c) ``quilt, comforter''} &     
    \footnotesize{d) ``window screen''}
  \end{tabular}
  \caption{ImageNet based BoS for a) bedroom image. Object recognition channels 
    for b) ``day bed'' c) ``comforter'' and d) ``window screen.''}
\label{fig:BoS_example}
\end{figure}

Throughout this work, we use ImageNet classes as $\cal V$ and object 
recognition CNNs to estimate the $\pi_i$. For efficient BoS extraction, the
CNN is implemented as a fully convolutional network, generating the
BoS with a single forward pass per image. This requires changing fully
connected into 1x1 convolutional layers. The receptive field of a fully
convolutional CNN can be altered by reshaping the size of the input image.
E.g. for 512x512 images, the fully convolutional implementation 
of~\cite{krizhevsky12} extracts SMNs from 128x128 pixel patches 
32 pixels apart. Figure~\ref{fig:BoS_example} illustrates
the high quality of the resulting semantics.
Recognizers of the ``bed'', ``window'' and ``quilt'' objects
exhibit are highly active in the regions where they appear in a
bedroom scene.

\section{Semantic Embedding}

To design the invariant embedding $\cal E$ we rely on the Fisher 
vector~(FV)~\cite{perronnin10,sanchez13}. In this section, we review the
FV and discuss its computation using the EM
algorithm. 
%We next discuss the design of the invariant embedding $\cal E$ using the Fisher vectors approach.

%Spatial average or max operations used in~\cite{kwitt12,feifei-obank14} omit much of the discriminative information necessary for BoS image classification. A sophisticated non-linear embedding such as a Fisher vector~(FV)~\cite{jakkola99_full,perronnin10} is, in fact, likely to be much better in any Bag-of-features recognition paradigm. For our ImageNET based SMNs, therefore, we explore the design of an appropriate Fisher vector embedding in the following sections. 

%\subsection{Fisher Vectors and EM}

%We start by a discussion of FVs and their computation using the EM algorithm.

\subsection{Fisher Vectors}

Images are frequently represented by a bag of 
descriptors ${\cal D} = \{\pi_1, \ldots, \pi_n\}$ sampled 
independently from some generative model $p(\pi; \theta)$.  An embedding
is used to map this representation into a fixed-length vector suitable for
classification. A popular 
mapping is the gradient (with respect to $\theta$) of the 
log-likelihood $\nabla_\theta L(\theta) = 
\frac{\partial}{\partial \theta}\log p({\cal D}; \theta)$ evaluated at a 
{\it background model\/} $\theta^b$. This is known as 
the {\it Fisher score\/} of $\theta$.
% The score encodes rich statistics of feature distribution within an image and serves as a good image representation for high level tasks such as matching, retrieval and classification. 
This gradient vector is often normalized by ${\cal F\/}^{-\frac{1}{2}} 
\nabla_\theta L(\theta)$, where ${\cal F\/}^{-\frac{1}{2}}$ is 
the square root of the Fisher information matrix ${\cal F\/}$ of 
$p(\pi; \theta)$. This is the FV of $\cal D$~\cite{jakkola99,perronnin10}. 

Since, for independent sampling, $\log p({\cal D}; \theta)$
is a sum of the log-likelihoods $\log p(\pi_i; \theta)$,
the FV is a vector of pooling operators, whose strength
depends on the expressiveness of the generative model $p(\pi; \theta)$. 
%An FV derived from a sophisticated model can capture higher 
%order trends of the feature distribution. 
The FV based on a large Gaussian mixture model~(GMM) is 
known to be a strong descriptor of image 
context~\cite{perronnin07,sanchez13}. However,
for models like GMMs or hidden Markov models, the FV can have various 
implementations of very different complexity and deriving an efficient
implementation is not always easy. We next show that Fisher scores can be 
trivially obtained using a single step of the expectation maximization~(EM) 
algorithm commonly used to learn such models. This unifies the EM
and FV computations, enabling the use of many efficient implementations 
previously uncovered in the EM literature to implement FVs.
 
%While the Fisher vector is frequently used with a
%Gaussian mixture model (GMM)~\cite{perronnin10,sanchez13}, 
%any generative model $p(x; \theta)$ can be used. However, the information matrix is not always easy to compute. When this is case, it 
%is common to rely on the simpler representation of ${\cal I}$ by
%the score $\nabla_\theta L(\theta)$. This is, for example, the case 
%with the sparse coded gradient vectors in~\cite{liu14}. 
%We next show that, for models with hidden 
%variables, the Fisher score can be obtained trivially from the steps of the 
%expectation maximization~(EM) algorithm commonly used to learn
%such models.  

\subsection{Fisher Scores from EM}
\label{sec:EM}
Consider the log-likelihood of ${\cal D}$
under a latent-variable model 
$\log p({\cal D}; \theta) = \log \int p({\cal D}, z; \theta) dz$
of hidden variable $z$. 
Since the left-hand side is independent of the hidden variable,
%The statistical approach known as ``slicing'' [??] can be used to derive
this can be written in an alternate form~\cite{DLR}
\begin{eqnarray}
\label{eq:ELBO}
%\lefteqn{
&&\log p({\cal D}; \theta) 
= \log p({\cal D}, z; \theta) - \log 
  p(z | {\cal D}; \theta) 
  \nonumber\\
&=& \int q(z) \log p({\cal D}, z; \theta) dz - 
    \int q(z) \log p(z | {\cal D}; \theta) dz
    \nonumber\\
&=& \int q(z) \log p({\cal D}, z; \theta) dz - 
    \int  q(z) \log  q(z) dz \nonumber\\
    &&+\int q(z) \log \frac{q(z)}{p(z | {\cal D}; \theta)} dz
  \nonumber\\
&=& Q(q; \theta) + H(q) + KL(q || p; \theta)
\end{eqnarray}
where $Q(q; \theta)$ is the ``Q'' function of the EM algorithm, 
$q(z)$ a general probability distribution, $H(q)$ its differential 
entropy and $KL(q||p; \theta)$ the Kullback Liebler divergence 
between the posterior $p(z | {\cal D}; \theta)$ and $q(z)$.
It follows that
\begin{eqnarray}
\label{eq:em1}
\frac{\partial}{\partial \theta} \log p({\cal D}; \theta) &=&
 \frac{\partial}{\partial \theta} Q(q; \theta) + 
\frac{\partial}{\partial \theta} KL(q || p; \theta)
\end{eqnarray}
where
\begin{eqnarray}
\frac{\partial}{\partial \theta} KL(q || p; \theta) &=&
-\int \frac{q(z)}{p(z | {\cal D}; \theta)} 
\frac{\partial}{\partial \theta} p(z | {\cal D}; \theta) dz.
\end{eqnarray} 
Each iteration of the EM algorithm chooses the $q$ distribution  
$q(z) = p(z | {\cal D}; \theta^b)$, where $\theta^b$
is a reference parameter vector (parameter estimates 
from previous iteration). In this case,
\begin{eqnarray}
  Q(q; \theta) &= \int p(z | {\cal D}; \theta^b) \log p({\cal D},z; \theta) dz \\
  &= E_{z | {\cal D}; \theta^b} [\log p({\cal D},z; \theta)]
\end{eqnarray}
and
\begin{eqnarray*}
  \left. \frac{\partial}{\partial \theta} KL(q || p; \theta)
  \right|_{\theta=\theta^b} &=& \left.
-\int \frac{p(z | {\cal D}; \theta^b)}{p(z | {\cal D}; \theta^b)} 
\frac{\partial}{\partial \theta} p(z | {\cal D}; 
\theta)\right|_{\theta=\theta^b} dz \\
&=& \left. - \frac{\partial}{\partial \theta} \int
 p(z | {\cal D}; \theta)\right|_{\theta=\theta^b} dz =0.
\end{eqnarray*}
It follows from~\eqref{eq:em1} that
\begin{eqnarray}
\left. \frac{\partial}{\partial \theta} 
  \log p({\cal D}; \theta)\right|_{\theta=\theta^b} &=&
  \left. \frac{\partial}{\partial \theta} Q(p(z | {\cal D}; \theta^b);
                                                   \theta).
  \right|_{\theta=\theta^b}
\end{eqnarray}

% An alternative approach to the same result, requires analysis of the KL divergence between $q = p(z|{\cal D}; \theta^b)$ and $p(z|{\cal D}; \theta)$. By rearranging terms in~(\ref{eq:ELBO}) it can be written as, 
% \begin{eqnarray}
% KL(\theta^b; \theta) = \log p({\cal D}; \theta) 
% - Q(\theta^b; \theta) - H_q(\theta^b) \nonumber
% \end{eqnarray}
% For models that allow exact inference, this expression is mathematically tractable and continuously differentiable with respect to $\theta$. The divergence, therefore, smoothly reduces to 0 as $\theta$ approaches $\theta^b$ from any direction. The slope of a tangent to $KL(\theta^b; \theta)$ (it's derivative) at $\theta = \theta^b$, therefore, equals 0. It follows that 
% \begin{eqnarray}
% \left. \frac{\partial}{\partial \theta} KL(\theta^b; \theta)\right|_{\theta=\theta^b} = 0\nonumber\\ 
% \left.\frac{\partial}{\partial \theta} \log p({\cal D}; \theta)\right|_{\theta^b} - \left.\frac{\partial}{\partial \theta} Q(p;\theta)\right|_{\theta^b} = 0 \nonumber\\
% \left.\frac{\partial}{\partial \theta} \log p({\cal D}; \theta)\right|_{\theta^b} = \left.\frac{\partial}{\partial \theta} Q(p;\theta)\right|_{\theta^b} \nonumber
%\end{eqnarray}

%In summary, the Fisher score $\nabla_\theta L(\theta)|_{\{\theta = \theta^b\}}$
%of background model $\theta^b$ is the gradient of 
%the Q-function of EM evaluated at reference model $\theta^b$. 
In summary, the Fisher score $\nabla_\theta L(\theta)|_{\{\theta = \theta^b\}}$ of 
background model $\theta^b$ is the gradient of 
the Q-function of EM evaluated at reference model $\theta^b$.
The computation of the Fisher score thus simplifies into the two steps of
EM. First, the E step computes the Q function 
$Q(p(z|{\cal D}; \theta^b); \theta)$ at the reference $\theta^b$. Second,
the M-step 
evaluates the gradient Q with respect to 
$\theta$ at $\theta = \theta^b$.
%This is exactly the computation performed in the M-step
%of the EM algorithm (where $\theta^b$ is the model obtained in the
%previous iteration).
Since latent variable models are learned with EM, efficient
implementations of these steps are usually already available
in the literature, e.g. the Baum-Welch algorithm
used to learn hidden Markov models~\cite{rabiner}.
Hence, the connection to EM makes the derivation  of the Fisher score trivial for most models of interest.

\section{Semantic Fisher vectors}
\label{sec:semE}
In this section, we discuss the encoding of the Image BoS into semantic FVs. 
%In vision, Fisher vectors are often derived using a Gaussian mixture model~(GMM)~\cite{perronnin10,sanchez13}. 

\subsection{Gaussian Mixture FVs}
\label{sec:gmmFV}
The most popular model in the FV literature is the 
GMM of diagonal
covariance~\cite{perronnin07,perronnin10,sanchez13},
here denoted  the variance-GMM. Under this generative model,
a mixture component $z_i$ is first sampled from a hidden variable $z$ of
categorical distribution $p(z = k) = w_k$. A descriptor $\pi_i$ is then
sampled from the Gaussian component
$p(\pi| z = k) \sim G(\pi, \mu_k, \sigma_k)$ 
of mean $\mu_k$ and variance $\sigma_k$, which is a diagonal matrix. 
Both hidden and observed variables are sampled independently.
The Q function is
\begin{eqnarray}
  Q(p(z|{\cal D};\theta^b); \theta) =   
  \sum\nolimits_i  E_{z_i | \pi_i; \theta^b} [\log p(\pi_i,z_i; \theta)] \nonumber\\
  = \sum\nolimits_i  E_{z_i | \pi_i; \theta^b}
  \left[\sum\nolimits_k I(z_i,k) \log p(\pi_i,k; \theta)\right] \nonumber \\
  =    \sum\nolimits_{i,k}  p(k | \pi_i; \theta^b)
  \log p(\pi_i | z_i = k; \theta) w_k
\end{eqnarray}
%\begin{eqnarray}
%  Q(p(z|{\cal D};\theta^b); \theta)
%  &=& \sum\nolimits_i  E_{z_i | x_i; \theta^b} \left[\sum\nolimits_k I(z_i,k) \log p(x_i,k; \theta)\right] \nonumber \\
%  &=&    \sum\nolimits_{i,k}  h_{ik} \log p(x_i | z_i = k; \theta) w_k
%\end{eqnarray}
where $I(.)$ is the indicator function. The probabilities
$p(k | \pi_i; \theta^b)$ are the only quantities computed in the E-step. 
The M-step then computes the gradient with respect to parameters 
$\theta = \{\mu_k, \sigma_k\}$ 
\begin{equation}
\label{eq:grad_mean}
{\cal G\/}_{\mu_k^d}({\cal I\/}) = 
\frac{\partial }{\partial \mu_k^d} Q =
\sum\nolimits_i p(k|\pi_i) 
\left(\frac{\pi_i^d - \mu_k^d}{(\sigma_k^d)^2}\right)
\end{equation}
\begin{equation}
\label{eq:grad_variance}
{\cal G\/}_{\sigma_k^d}({\cal I\/}) = 
\frac{\partial }{\partial \sigma_k^d} Q
= \sum\nolimits_i p(k|\pi_i) 
\left[\frac{(\pi_i^d - \mu_k^d)^2}{(\sigma_k^d)^3} - \frac{1}{\sigma_k^d}\right],
\end{equation}
where $Q$ indicates the log-likelihood of the image and $\pi_i^d$ is
the $d^{th}$ entry of vector $\pi_i$.

These are also the components of the Fisher score, when
evaluated using a reference model $\theta^b = \{\mu_k^b, \sigma_k^b\}$ learned (with EM) from all training data.
%To compute the Fisher vectors, scores in~(\ref{eq:grad_mean}) and~(\ref{eq:grad_variance}) are often scaled by an approximate Fisher information matrix, as detailed in~\cite{sanchez13}.  
The FV is obtained by scaling the gradient vectors by an approximate Fisher information matrix, as detailed in~\cite{sanchez13}. This leads to the following mean and variance components of the GMM-FV 
\begin{equation}
\label{eq:fv_mean}
{\cal V\/}_{\mu_k}({\cal I\/}) = 
\frac{1}{n\sqrt{w_k}}\sum\nolimits_i p(k|\pi_i) 
\left(\frac{\pi_i - \mu_k}{\sigma_k}\right)
\end{equation}
\begin{equation}
\label{eq:fv_variance}
{\cal V\/}_{\sigma_k}({\cal I\/}) = 
\frac{1}{n\sqrt{2w_k}}\sum\nolimits_i p(k|\pi_i) 
\left[\frac{(\pi_i - \mu_k)^2}{\sigma_k^2} - 1\right].
\end{equation}
For a single Gaussian component of zero mean, \eqref{eq:fv_mean}
reduces to the average pooling operator. For mixtures of many components, \eqref{eq:fv_mean} implements a pooling operator per component, restricting each operator to descriptors of large probability $p(k|\pi_i)$ under the component.
The FV can also implement other pooling operations, e.g. capturing higher order statistics as in \eqref{eq:fv_variance}.
Many variations of the GMM-FV have been proposed to enable discriminative
learning~\cite{maaten11}, spatial feature encoding~\cite{krapac11} or non-iid
mixture modeling~\cite{cinbis12}. However, for  low-level features
and large enough mixtures, the classical FV of~\eqref{eq:fv_mean}
and~\eqref{eq:fv_variance} is still considered state-of-the-art.  
%This is determined by the background probability model.
%descriptors
%these mean and variance scores usually capture complimentary discriminative information, useful for image classification~\cite{perronnin10}. 
%Yet, 
%FVs computed from CNN features only use the mean gradients 
%similar to~(\ref{eq:grad_mean}), ignoring second-order 
%statistics~\cite{gong14,dixit15}. In the experimental section, we show that the variance statistics of CNN features perform poorly compared to the 
%mean gradients. %and can actually degrade performance. 
%This is perhaps due 
%to the inability of the variance-GMM to accurately model data 
%in high dimensions. We test this hypothesis by considering a 
%model better suited for this task.

\subsection{Dirichlet Mixture FVs}
\label{sec:dmmFV}

The variance-GMM is a default model for low-level visual
descriptors~\cite{nuno00,csurka04,sanchez13,jegou10}. However, SMNs,
which inhabit a probability simplex, are more naturally modeled by the
Dirichlet mixture~(DMM).  
This follows from the fact that the Dirichlet distribution is the most
popular model for probability vectors~\cite{minka00}. For example, it is
widely used for text modeling~\cite{blei03}, as a prior of the latent
Dirichlet allocation model, and for SIFT based image
categorization~\cite{feifei-lda05,cinbis12}. The DMM was
previously  used to model ``theme'' based SMNs in~\cite{rasiwasia11}. It is
defined as
\begin{eqnarray}
\label{eq:dir-pdf}
P(\pi| \{\alpha_k, w_k\}_{k=1}^K) &=& 
\frac{1}{Z(\alpha_k)}e^{\sum_l (\alpha_{kl} - 1)\log \pi_{l}}.
\end{eqnarray}
where $\alpha_k$ is the Dirichlet parameter of the $k^{th}$ mixture component 
and $w_k$ denotes the mixture weight. $Z(\alpha_k)$ is the normalizing constant 
$\frac{\gamma\left(\sum_l \alpha_{kl} \right)}{\prod_l \gamma(\alpha_{kl})}$, 
where $\gamma(x) = \int_0^{\infty} x^{t-1}e^{-x} dx$ is the Gamma function. The 
generative process is as follows. A mixture component $z$ is sampled from a 
categorical distribution $p(z = k) = w_k$. An observation $\pi$ is then 
sampled from the selected Dirichlet component $P(\pi | \alpha_k)$. This makes
the observation $\pi$ a multinomial distribution that 
resides on the probability simplex. 

The  EM algorithm for DMM learning has $Q$ function 
\begin{eqnarray}
  Q(p(z|{\cal D};\alpha^b); \alpha) =  \sum_{i,k}  
  h_{ik} \left(\sum_l \alpha_{kl}\log \pi_{l} - \log Z(\alpha_k)\right)
\end{eqnarray}
where $h_{ik}$  is the posterior probability $p(k | \pi_i; \theta_b)$ of 
the sample $\pi_i$ being under the $k^{th}$ components and we ignore
terms that do not depend on the $\alpha$ parameters\footnote{Gradients w.r.t 
  mixture weights $w_k$ are less informative than w.r.t other parameters
  and ignored in the
FV literature~\cite{perronnin07,perronnin10,sanchez13}.}. 
The expression for the Fisher scores ${\cal G}_{\alpha_k}({\cal I}) =
\frac{\partial L(\theta)}{\partial \alpha_k}$
of a DMM is
\begin{multline}
\label{eq:grad_alpha}
{\cal G}_{\alpha_k}({\cal I}) = \frac{1}{n}\sum_{i=1}^N h_{ik} \left(\log \pi_i
  - \psi(\alpha_k) + \psi(\sum_l \alpha_{kl}) \right),
\end{multline}
where $\psi(x) = \frac{\partial \gamma(x)}{\partial x}$. As usual
in the FV literature~\cite{perronnin10}, 
we approximate the Fisher information ${\cal F\/}$ by the
component-wise block diagonal matrix
\begin{equation}
\label{eq:FIM_dmm}
\begin{split}
\left({\cal F}_k\right)_{lm} &= E \left[ -\frac{\partial^2 \log P(\pi | \{\alpha_k, w_k\}_{k=1}^K )}{\partial \alpha_{kl} \partial \alpha_{km}} \right]\\
&\approx w_k \left(\psi'(\alpha_{kl})\delta(l, m) - \psi'(\sum_l\alpha_{kl}))\right)
\end{split}
\end{equation}
where $\delta(l, m) = 1$ if $l = m$. 
%A complete derivation of ${\cal F\/}$ is given in Section~\ref{sec:directFV_supp} of the supplement. 
The DMM Fisher vector for image $I$ is finally obtained 
from~(\ref{eq:grad_alpha}) and~(\ref{eq:FIM_dmm}) as 
${\cal F}_k^{-1/2}{\cal G}_{\alpha_k}({\cal I\/})$.

\subsection{The logSMN-FV}

To understand the benefits and limitations of the GMM-FV and
DMM-FV it helps to investigate their relationships.
Consider the application of the two FVs to the set of SMNs 
$\{\pi_1, \ldots, \pi_n\}$ extracted from image $\cal I$.
In both cases, the FV can be written as
\begin{equation}
  \label{eq:FV_general}
  {\cal V}_{\theta_k}({\cal I}) 
  = \frac{1}{n}\sum_{i=1}^N p(k | \pi_i) \gamma(\theta_k) \left(\nu(\pi_i) -
  \xi(\theta_k)\right)
\end{equation}
where $\gamma(.), \nu(.),$ and $\xi(.)$ are defined in
Table~\ref{tab:relations}.
This is a pooling mechanism that combines four operations: 
$p(k | \pi_i)$ {\it assigns\/} the SMNs $\pi_i$ to the components $k$,
$\nu(.)$ {\it embeds\/} each SMN into the space where pooling takes place,
$\xi(.)$ defines a {\it centroid\/} with respect to which the residuals
$\nu(\pi_i) - \xi(\theta_k)$ are computed, and $\gamma(\theta_k)$
{\it scales\/} or normalizes that residual. 

\begin{figure}[t]\RawFloats
  \centering
    \captionof{table}{Parameters of \eqref{eq:FV_general} for the 
      GMM-FV and DMM-FV. ${\cal F}_k$ is given by \eqref{eq:FIM_dmm},
      $h_k(\pi; \mu, \Sigma, w)$ by \eqref{eq:hk} and 
      $q_k(\pi; \alpha, w)$ by \eqref{eq:qk}.
    }
    \label{tab:relations}
    \small
    \setlength{\tabcolsep}{2pt}
    \begin{tabular}{|l|c|c|}
      \hline
      & GMM-FV & DMM-FV \\
      \hline
      $\theta_k$ & $\mu_k$, $\Sigma_k = \sigma_k I$ & $\alpha_k$ \\
      \hline 
      $\nu(\pi_i)$ & $\pi_i$ & $\log(\pi_i)$\\
      \hline
      $\xi(\theta_k)$ & $\mu_k$ & $f(\alpha_k) = \psi(\alpha_k) -
                                  \psi(\sum_l \alpha_{kl})$\\
      \hline
      $\gamma(\theta_k)$ & $\frac{1}{\sqrt{w_k} \sigma_k}$ & ${\cal F}_k^{-1/2}$\\
      \hline
      $p(k|\pi)$ & $h_k(\pi; \mu, \Sigma, w)$
               & $q_k(\pi; \alpha, w)$ \\
      \hline
    \end{tabular}
\end{figure}

There are three main differences
between the FVs. First, while the GMM-FV lacks an embedding, the
DMM-FV uses $\nu(.) = \log(.)$. Second, while the
GMM-FV has independent parameters to define centroids ($\mu_k$) and
scaling ($\sigma_k$), the parameters of the DMM-FV are coupled, since the
centroids $f(\alpha_k)$ and the scaling parameter ${\cal F}_k^{-1/2}$
are both determined by the DMM parameters $\alpha_k$.
Finally, the two FVs differ in the assignments $p(k | \pi_i)$ and
centroids $\xi(\theta_k)$. However, the centroids are closely related.
Assuming a background mixture model learned from a training set
$\{\pi_1^b, \ldots, \pi_N^b\}$ they are the parameters that
set \eqref{eq:fv_mean} and \eqref{eq:grad_alpha} to zero upon
convergence of EM. This leads to the expressions
\begin{eqnarray}
  \label{eq:muk}
  \mu_k & = & \frac{\sum_i p(k|\pi_i^b) \pi_i^b}{\sum_i p(k|\pi_i^b)} \\
  \label{eq:fk}
f(\alpha_k) & = & \frac{\sum_i p(k|\pi_i^b) \log \pi_i^b}{\sum_i p(k|\pi_i^b)}.
\end{eqnarray}
The differences in the assignments are also mostly of detail, since 
\begin{eqnarray}
  h_k(\pi; \mu, \Sigma, w) &=& \frac{w_k e^{||\pi - \mu_k||_{\Sigma_k}}}
                               {\sum_j w_j e^{||\pi - \mu_j||_{\Sigma_j}}}
  \label{eq:hk} \\
  q_k(\pi; \alpha, w) &=& \frac{w_k e^{(\alpha_k-1)^{T} \log\pi}}
                          {\sum_j w_j e^{(\alpha_j-1)^{T} \log\pi}} 
                          \label{eq:qk}
\end{eqnarray}
are both softmax type non-linearities. For both assignments and centroids,
the most significant difference is the use of the $\log \pi$ embedding
in the DMM-FV.

In summary, the two FVs differ mostly in the use of the $\log \pi$ embedding
by the DMM-FV and the greater modeling flexibility of the GMM-FV, due
to the availability of independent localization (centroid) $\mu_k$
and scale $\sigma_k$ parameters. This suggests the possibility of
combining the strengths of the two FVs by applying the GMM-FV {\it after\/}
this embedding. We refer to this as the logSMN-FV
\begin{equation}
\label{eq:fvlog_mean}
{\cal V\/}_{\mu_k}({\cal I\/}) = 
\frac{1}{n\sqrt{w_k}}\sum\nolimits_i p(k|\pi_i) 
\left(\frac{\log \pi_i - \mu_k}{\sigma_k}\right).
\end{equation}
Our experiments, see Section~\ref{sec:piFV} (Table \ref{tab:smnFVs}), show that 
this simple transformation leads to a large improvement in classification 
accuracy. 

\begin{figure}[t]
  \centering
  \begin{tabular}{cc}
    \includegraphics[width=.38\linewidth]{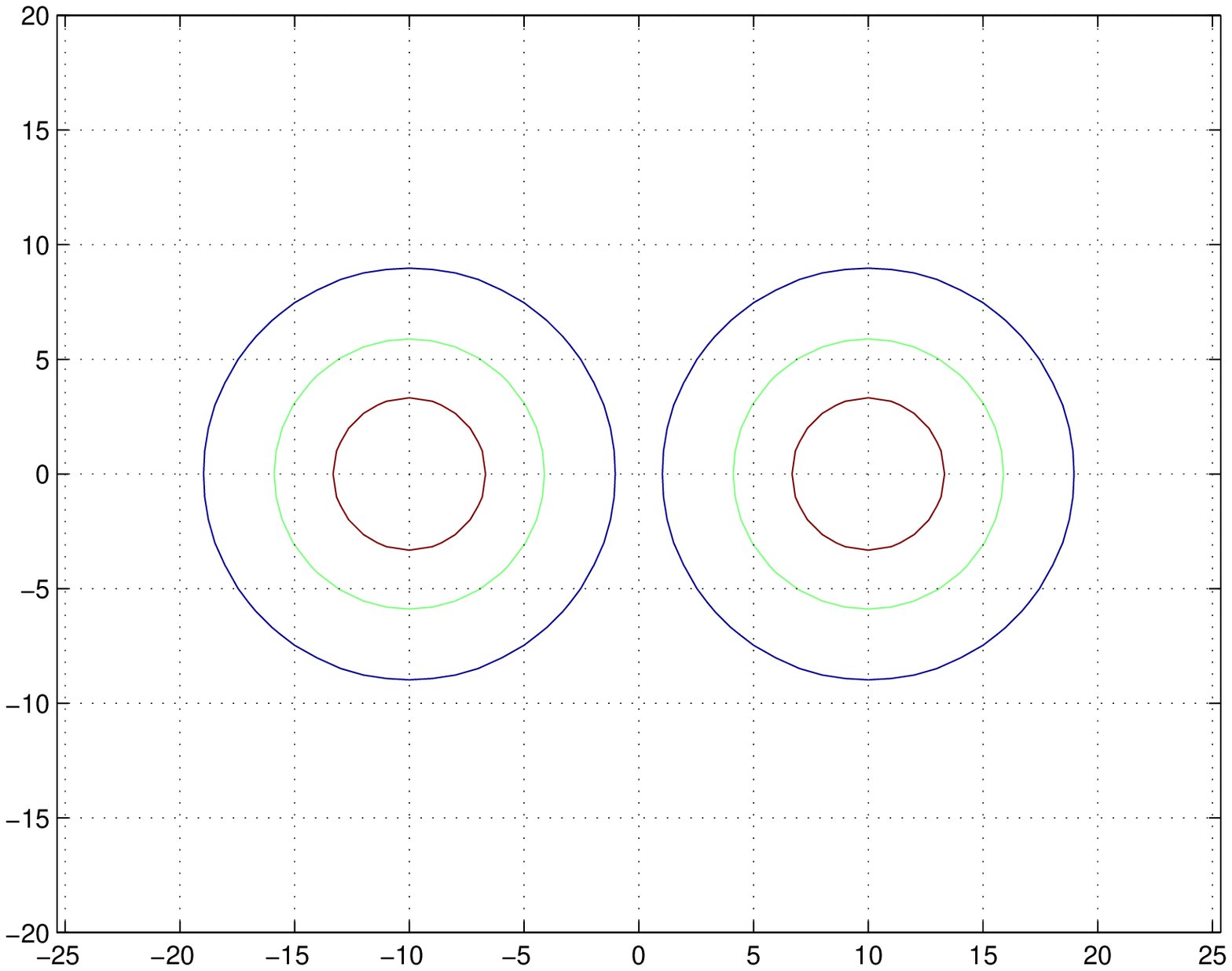} &
    \includegraphics[width=.38\linewidth]{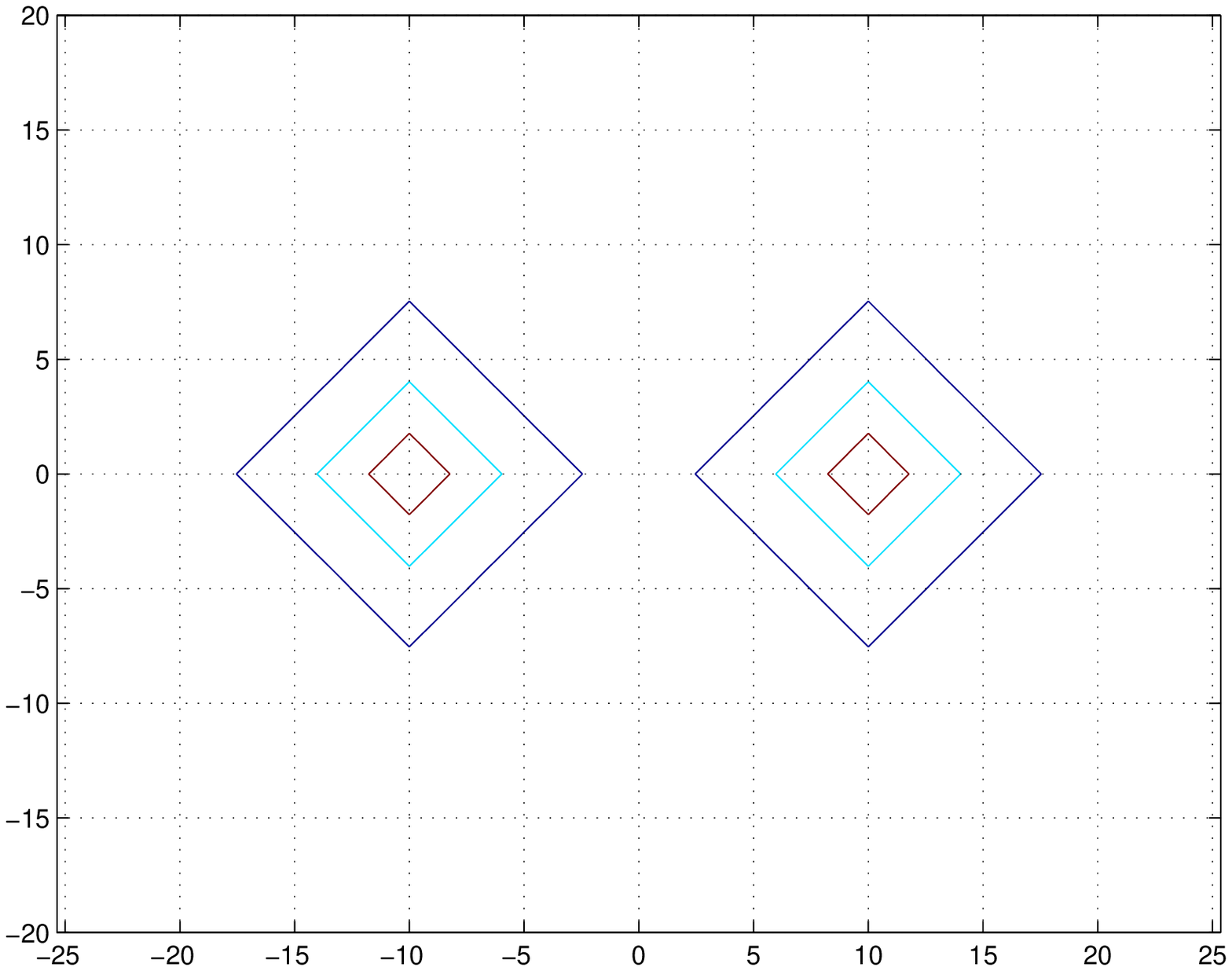} \\
    \footnotesize{a)} &     \footnotesize{b)}\\
    \includegraphics[width=.38\linewidth]{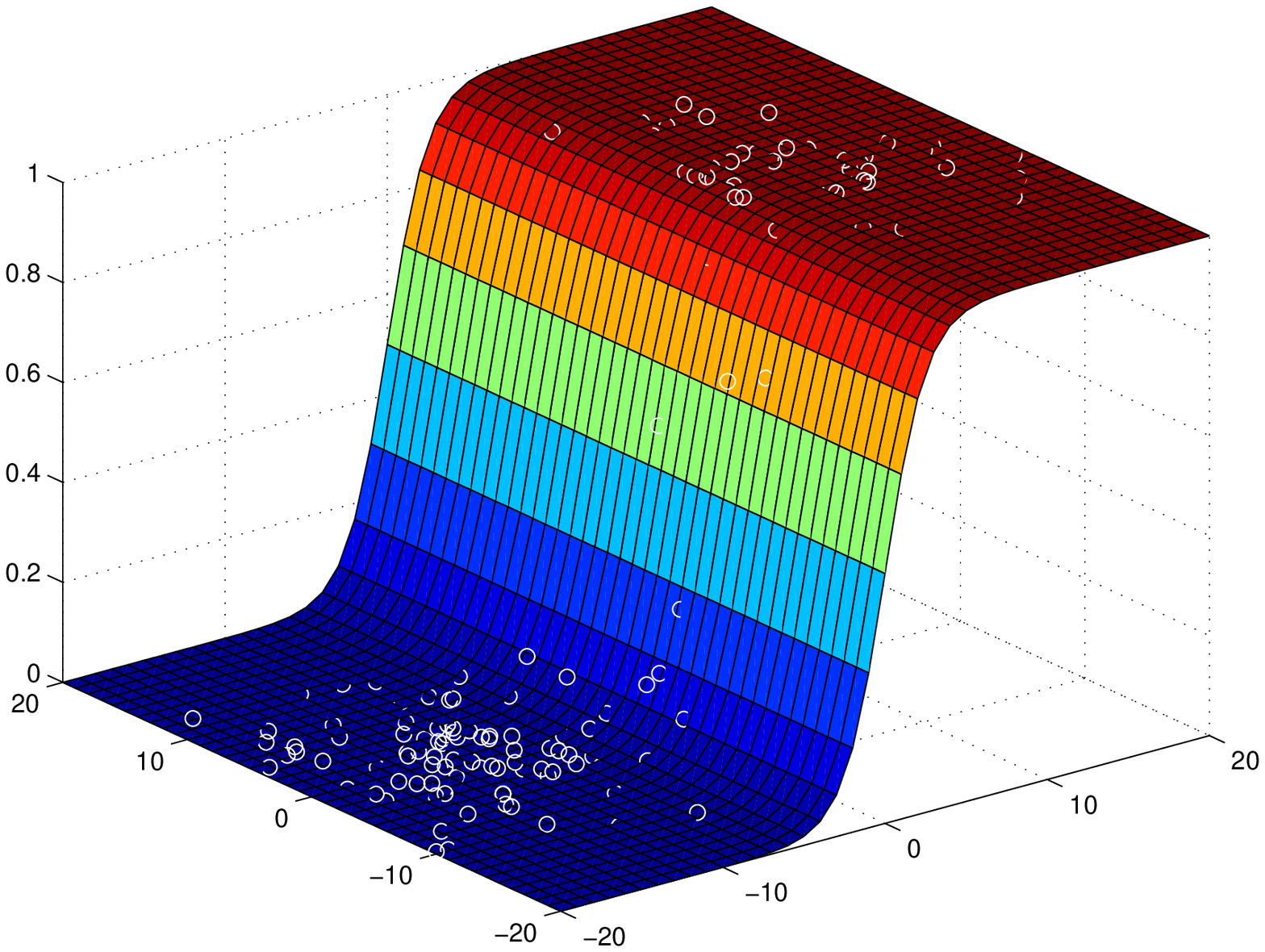} &
    \includegraphics[width=.38\linewidth]{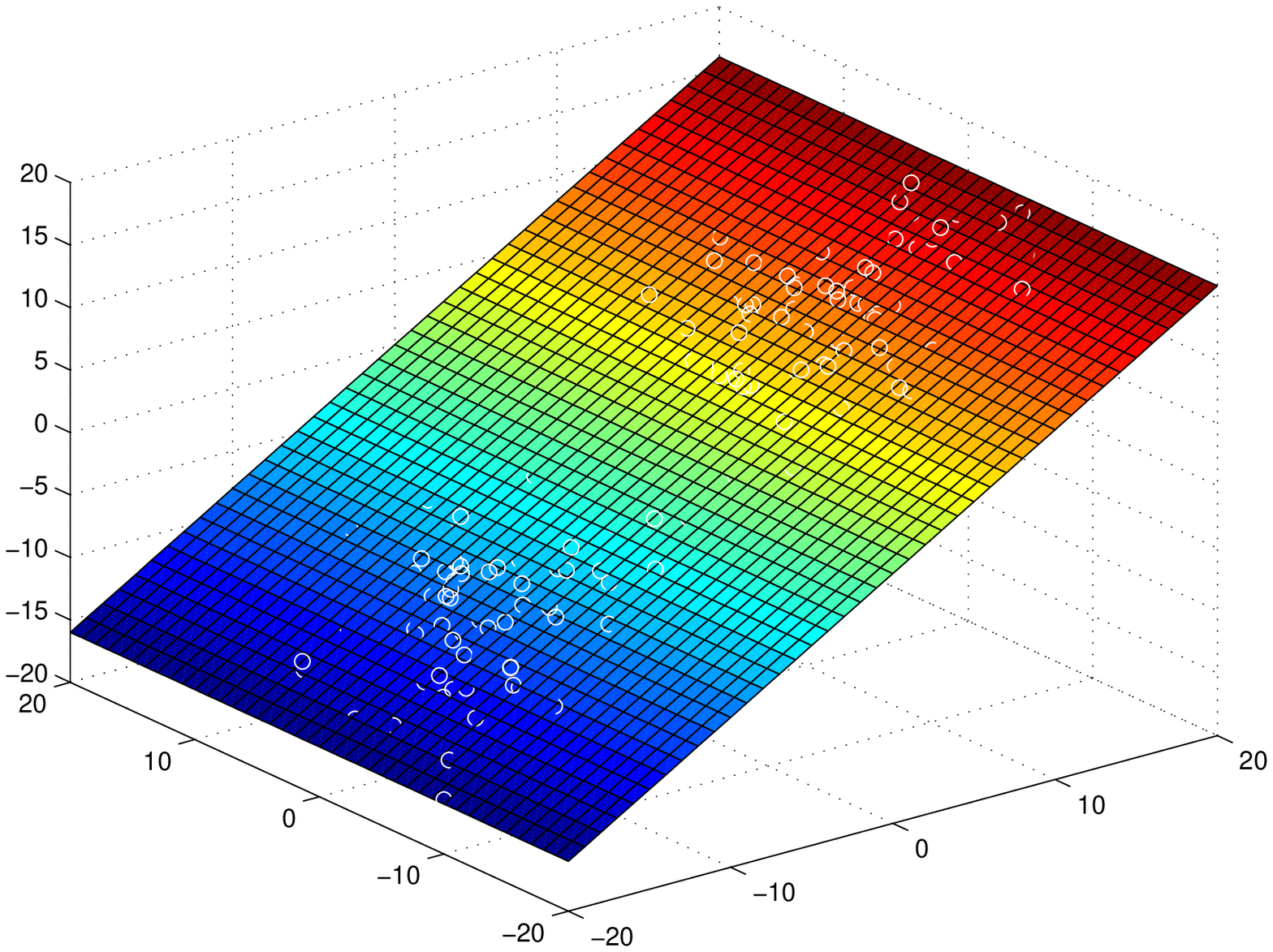} \\
    \footnotesize{c)} &     \footnotesize{d)}
  \end{tabular}
  \caption{Top: Two classifiers in an Euclidean space $\cal X,$
   a) $L_2$ and b) $L_1$ merics. Bottom: c) projection of
    a sample from a) into the semantic space $\cal S$ (only $P(y=1|x)$ shown).
    d) natural parameter space mapping of c).
  }
  \label{fig:natural}
\end{figure}

\subsection{FVs in Natural Parameter Space}
%\label{sec:manifold}
%While the DMM is a natural model for SMNs, our experiments show
%that the DMM FV does not result in an effective
%scene classifier (see Section~\ref{sec:expMaps}).
\label{sec:npBoS}

% A drawback of the variance-GMM FV is its use of direct mixture modeling
% in the probability simplex.
% %Learning efficient models in this space is
% %not easy due its non-Euclidean nature.
% In general, the difficulty of
% modeling in a space $\cal X$ depends on its topology. Most machine learning
% assumes Euclidean  spaces, e.g. where the natural
% measure of distance between samples $x_i \in \cal X$ is a metric. This is
% not the case of a probability simplex, which has a non-metric
% Kullback-Liebler divergence as its natural distance measure, making
% modeling quite difficult.  
% %In this section, we provide some explanations for the good performance
% %of the log mapping.

%The poor performance of the SMN-FV in Table~\ref{tab:smnFVs}
The gains of the log embedding can be explained by the non-Euclidean nature 
of the probability simplex.
% While GMM FVs perform reasonably well with low-level feature spaces such as SIFT, HoG etc., their failure with ImageNet SMNs can be attributed to a very non-Euclidean nature of the space of probability vectors.
% % which makes all
% %learning very difficult in this space.
% %DMM is not
% %as effective as its standard application to a Gaussian mixture.
% %This can be seen in Table~\ref{tab:1-embeddings}, where the semantic
% %DMM-FV underperforms
% %all other methods on MIT Indoors, and
% In general, the difficulty of modeling
% on a data space $\cal X$ depends on its topology.
% Most machine learning assumes vector spaces
% with Euclidean structure, e.g. where the
% natural measure of distance between examples $x_i \in {\cal X}$ is
% a metric. This is not the case for the probability
% simplex, which has a non-metric Kullback-Leibler
% divergence as its natural distance measure,
% and makes model learning quite difficult.
% %A hint of this is the fact that,
% %while there is a large diversity of probability density models for
% %Euclidean data, the Dirichelet is usually adopted on the simplex.
For some insight on this, consider the two binary classification problems
of Figures~\ref{fig:natural} a) and b).
In a) the two classes are Gaussian, in b) Laplacian. Both problems have
class-conditional distributions 
$P(x|y) \propto \exp\{-d(x, \mu_y)\}$ where
$Y \in \{0,1\}$ is the class label and
$d(x, \mu) = ||x-\mu||_p$, with $p=1$ for Laplacian and $p=2$ for 
Gaussian. Figures~\ref{fig:natural} a)
and b) show the iso-contours of the probability distributions under
the two scenarios. Note that the two classifiers use
very different metrics.

The posterior distribution of class $Y=1$ is, in both cases,
\begin{equation}
  \pi(x) = P(y=1|x) = \sigma(d(x,\mu_0) - d(x,\mu_1))
\end{equation}
where $\sigma(v) = (1+e^{-v})^{-1}$ is the sigmoid.
Since this is very non-linear, the projection $x
\rightarrow (\pi(x), 1-\pi(x))$ of the samples $x_i$ into the
semantic space destroys the Euclidean structure of the original
spaces $\cal X$.  This is illustrated in c),
which shows the posterior surface and the projections $\pi(x_i)$ for
Gaussian $x_i$. In this space, the shortest path between two samples is not
a line. The sigmoid also makes the posterior surfaces of 
the two problems very similar. The surface of the 
Laplacian problem in b) is visually indistinguishable from c).
In summary, Euclidean classifiers with two very different
metrics transform the data into highly non-Euclidean semantic
spaces that are almost indistinguishable.  This reduces the
effectiveness of modeling probabilities directly with GMMs or DMMs, 
producing weak FV embeddings.

The problem can be avoided by noting that SMNs are the parameters of 
the multinomial, which is a member of the exponential family of distributions
\begin{equation}
  P_S(s; \pi) = h(s) g(\pi) \exp \left(\eta^T(\pi) T(s)\right),
\end{equation}
where $T(s)$ is denoted a sufficient statistic.
In this family, the re-parametrization $\nu = \eta(\pi)$
makes the (log) probability distribution {\it linear\/} in the sufficient
statistic
\begin{equation}
  P_S(s; \nu) = h(s) g(\eta^{-1}(\nu)) \exp \left(\nu^T T(s)\right).
\end{equation}
This is called the {\it natural parameterization\/} of the
distribution. Under this parametrization, the multinomial
log-likelihood of the BoS in~(\ref{eq:smn-exp-lkd})
yields a natural parameter vector $\nu_i = \eta(E\{s_i\})$
for each patch $x_i$, instead of a probability vector.
For the binary  semantics of Figure~\ref{fig:natural}, $\eta(.)$ is the logit
transform $\nu = \log \frac{\pi}{1-\pi}$.
This maps the high-nonlinear semantic space of Figure~\ref{fig:natural} c)
into the linear space of d), which preserves the Euclidean
structure of a) and b). Hence, while the variance-GMM is not well matched 
to the geometry of the probability simplex where $\pi$ is defined, it 
is a good model for distributions on the (Euclidean) natural parameter 
space defined by $\nu(\pi)$.

Similarly, for multiclass semantics, the mapping from multinomial 
to natural parameter space is a one-to-one transformation 
into a space with Euclidean structure. 
%In fact, it can now be implemented by 
%the PCA in~(\ref{eq:FV_PCA}) and
%the encoding operation in~(\ref{eq:FV_mean}).
% lternatively, one could design a Gaussian mixture FV that explicitly encodes covariance information. In the next section, we introduce such a descriptor.
In fact, the multinomial of parameter vector
$\pi = (\pi_1, \ldots, \pi_S)$
has three possible natural parametrization
\begin{eqnarray}
  \nu_k^{(1)} &=& \log \pi_k \label{eq:nu1}\\
  \nu_k^{(2)} &=& \log \pi_k + C \label{eq:nu2}\\
  \nu_k^{(3)} &=& \log \frac{\pi_k}{\pi_S} \label{eq:nu3}
\end{eqnarray}
where $\nu_k$ and $\pi_k$ are the $k^{th}$ entries of $\nu$ and $\pi$,
respectively. 
%The performance of these parametrizations is likely
%to depend on the implementation of the semantic classifier that generates the SMNs. For a discriminant classifier such as a CNN, $\nu^{(2)}$ will likely be the best transformation.  
The fact that logSMNs implement $\nu^{(1)}$ explains the good performance of 
the logSMN-FV. However, the existence of  two alternative embeddings 
raises the question of whether this is the best natural parameter space 
embedding for the BoS produced by a CNN. 
Note that, under $\nu^{(2)}$, $\pi_k = \frac{1}{C} e^{\nu_k}$
defines a probability vector if and only if $C = \sum_i e^{\nu_i}$. Hence,
the mapping from $\nu^{(2)}$ to $\pi$ is the softmax function commonly
implemented at the CNN output. This implies that CNNs learn to optimally
discriminate data in the natural parameter space defined by $\nu_k^{(2)}$
and, for CNN semantics, $\nu_k^{(2)}$ should enable better scene
classification.

\begin{figure}[t]
  \centering
  \begin{tabular}{cc}
    \includegraphics[width=.39\linewidth]{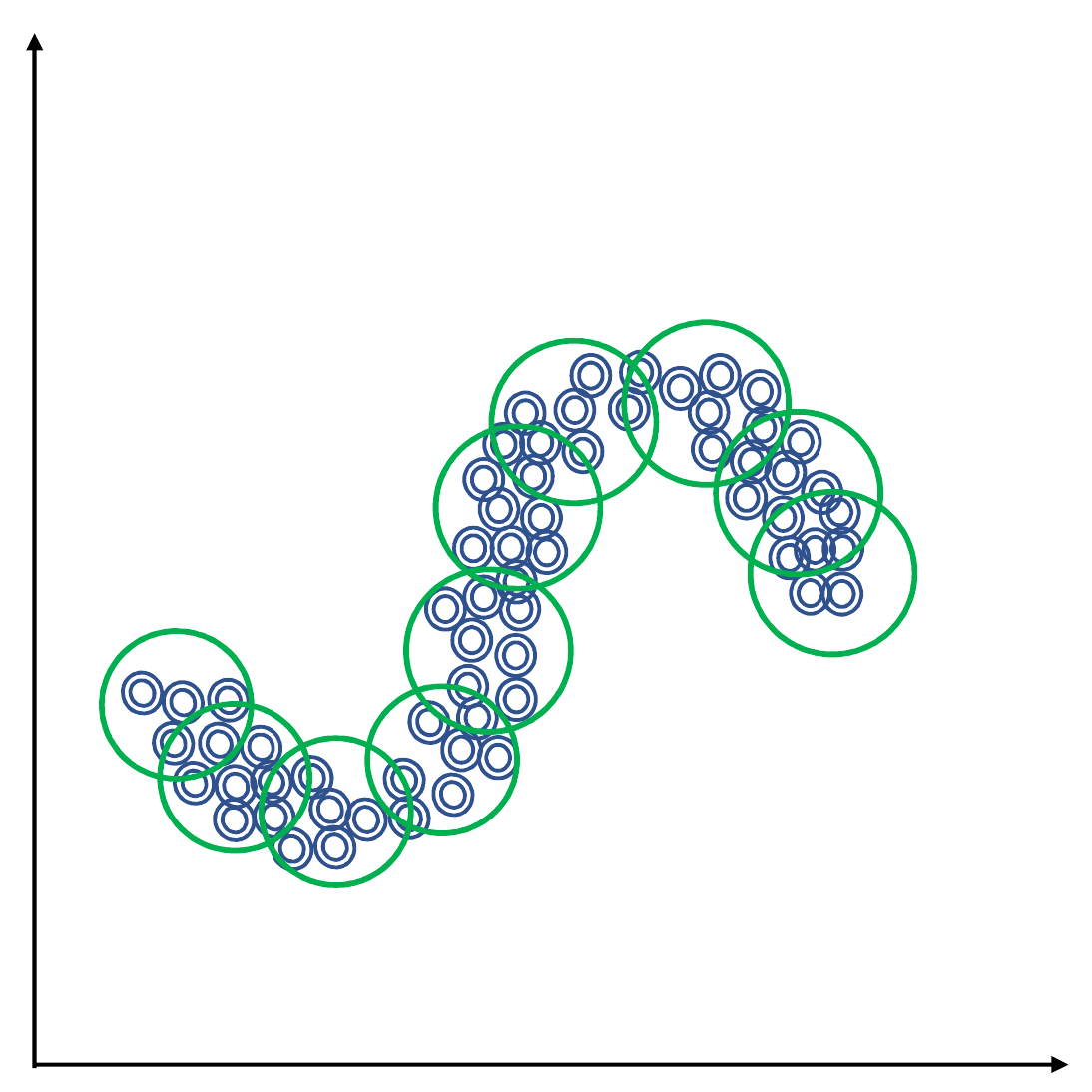} &
    \includegraphics[width=.39\linewidth]{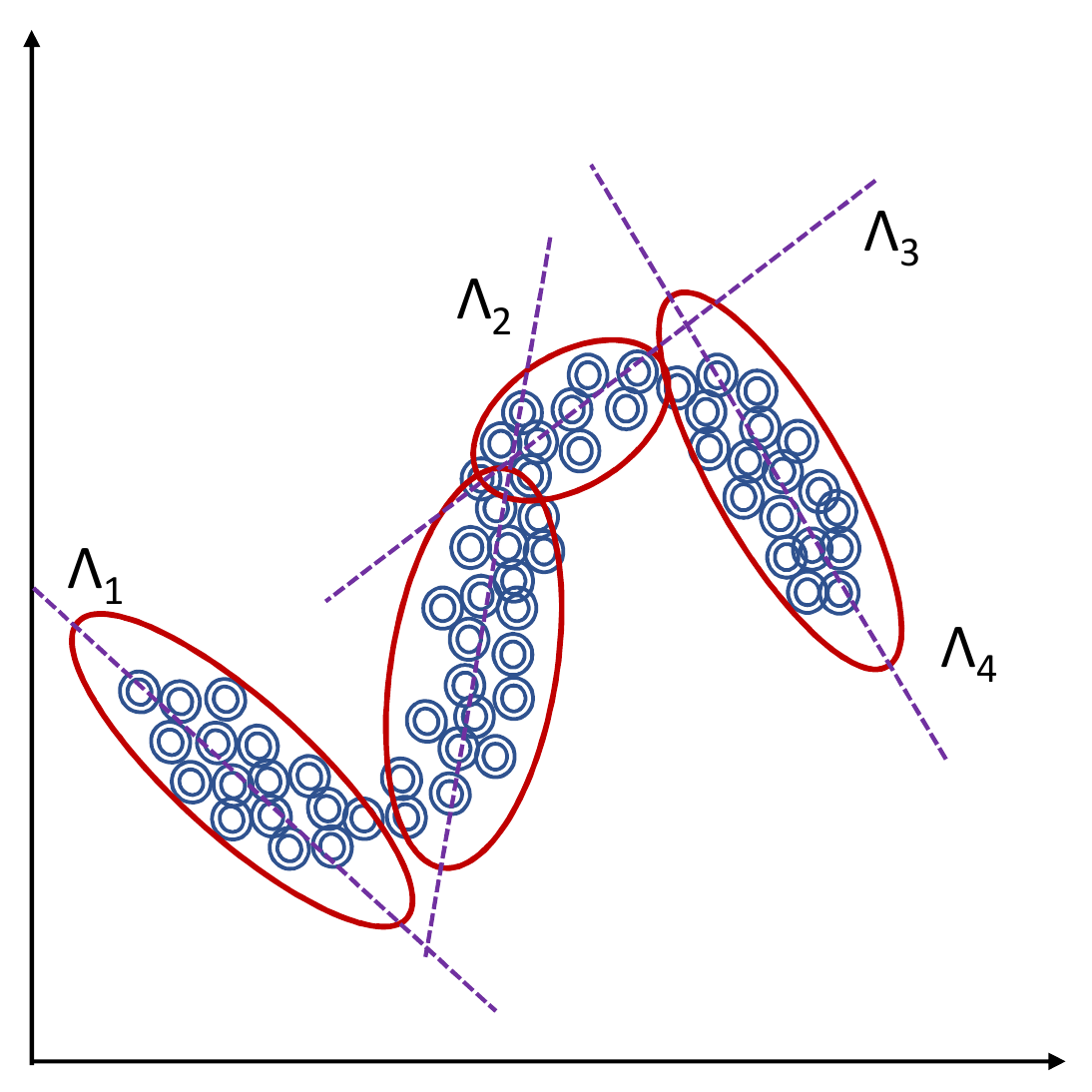} \\
    \footnotesize{a) variance GMM} &     \footnotesize{b)  MFA}\\
  \end{tabular}
  \caption{Modeling data on a manifold. The variance-GMM
    requires many Gaussians. By fitting locally linear subspaces,
    the MFA requires few Gaussians.}
\label{fig:MFA_example}
\end{figure}

\subsection{The MFA-FV}

The models introduced so far mostly disregard semantic feature covariance.
The Dirichlet mixture is, by design, incapable of modeling second order
statistics. As usual in the FV 
literature~\cite{perronnin07,perronnin10,sanchez13}, the GMM-based
FVs assume a diagonal covariance per mixture component.
% We refer to this as the variance-GMM. 
While standard for SIFT descriptors~\cite{sanchez13}, this
is not suitable for the much higher dimensional
CNN features, more likely to populate a low-dimensional manifold of the
ambient semantic space. As illustrated in Figure~\ref{fig:MFA_example}, the
variance-GMM requires many components to cover such a distribution. While a
full covariance GMM could be substantially more efficient, covariance
modeling is difficult in high dimensions. The data available for transfer 
learning is rarely sufficient to learn full covariances. 

In this work, we explore approximate covariance modeling using mixtures of
factor analyzers~(MFAs)~\cite{ghahramani97}. As illustrated in
Figure~\ref{fig:MFA_example}, the MFA approximates a non-linear data
manifold by a set of locally linear subspaces. Each mixture component
generates Gaussian data in a low dimensional latent space, which is then
projected linearly into the high dimensional observation space. This is a
low rank approximation of the full covariance Gaussian, which can be
learned with the small amounts of data available for transfer learning.
It generates high-dimensional covariance statistics that can be exploited
by a FV for better classification.

\subsubsection{MFA Fisher scores} 

A factor analyzer~(FA) models high dimensional observations 
$x \in \mathbb{R}^D$ in terms of latent ``factors'' 
$z \in \mathbb{R}^R$ defined on a low-dimensional subspace 
$R << D$~\cite{ghahramani97}. 
Specifically, $x = \Lambda z + \epsilon$,
%\begin{equation}
%\label{eq:FA}
%x = \Lambda z + \epsilon
%\end{equation}
where $\Lambda$ is the factor loading matrix and $\epsilon$ 
additive noise. Factors $z$ are distributed as $G(z, 0, I)$ and noise as 
$G(\epsilon, 0, \psi)$, where $\psi$ is a diagonal matrix. It can be 
shown that $x$ follows a Gaussian distribution $G(x,0, S)$ of 
covariance $S = \Lambda\Lambda^{T} + \psi$. Since this is a full covariance
matrix, the FA is better suited for high dimensional data than a Gaussian
of diagonal covariance.

The MFA extends the FA so as to allow a piece-wise linear 
approximation of a non-linear data manifold. It has two hidden variables:
a discrete variable $s$, $p(s = k) = w_k$, which 
determines the mixture assignments and a continuous latent variable 
$z \in \mathbb{R}^R$, $p(z | s=k) = G(z, 0, I)$, which is
a low dimensional projection of the observation variable 
$x \in \mathbb{R}^D$,
$p( x | z, s=k) = G(x, \Lambda_k z + \mu_k, \psi)$.
Hence, the $k^{th}$ MFA component is a FA of mean $\mu_k$ and subspace defined
by $\Lambda_k$. As illustrated in Figure~\ref{fig:MFA_example}, the MFA 
components approximate the distribution of $x$ by a set of sub-spaces.
The MFA can be learned with an EM algorithm of Q function
\begin{eqnarray*}
  \lefteqn{Q(\theta^b; \theta) =} & & \\
  &=& \sum_i  E_{z_i, s_i | x_i; \theta^b} \left[\sum\nolimits_k I(s_i,k) \log p(x_i,z_i, s_i = k; \theta)\right]  \\
  &=& \sum_{i,k}  h_{ik}
      E_{z_i|x_i; \theta^b} \Big[\log G(x_i, \Lambda_k z_i + \mu_k, \psi) \\ 
  &+& \log G(z_i, 0, I) + \log w_k\Big]
\nonumber
\end{eqnarray*}
where $h_{ik} = p(s_i = k | x_i; \theta^b)$. After some simplifications,
defining
\begin{eqnarray}
  S_k^b &=& \Lambda_k^b \Lambda_k^{b^{T}} + \psi^b \label{eq:mfaNet_cov}\\
  \beta_k^b &=& \Lambda_k^{b^{T}}\left(S_k^b\right)^{-1}, \label{eq:beta_k}
\end{eqnarray}
the E step reduces to computing 
\begin{eqnarray}
\label{eq:mfa_e1}
  h_{ik} &=&  p(k | x_i; \theta^b) \propto w_k^b G(x_i, \mu_k^b, S_k^b)\\
\label{eq:mfa_e2}
  E_{z_i|x_i; \theta^b}[z_i] &=& \beta^b_k(x_i - \mu_k^b)\\
  \label{eq:mfa_e3}
  E_{z_i|x_i; \theta^b}[z_i z_i^{T}] &=&
   \beta_k^b(x_i - \mu_k^b)(x_i - \mu_k^b)^{T}\beta_k^{b^{T}} \\
   &-& \left(\beta_k^b\Lambda_k^b - I\right).
\end{eqnarray}  
The M-step computes the Fisher scores of
$\theta = \{\mu_k^b, \Lambda_k^b\}$. After some algebraic manipulations, these
can be written as
% \begin{eqnarray}
%\label{eq:mfa_mean_1}
%{\cal G\/}_{\mu_k}({\cal I\/}) &=& \sum_{i=1}^n p(k|x_i; \theta^b) \psi^{b^{-1}}\left(x_i - \Lambda_k^b E\left[z_i; \theta^b\right] - \mu^b_k\right)\\
%\label{eq:mfa_lambda_1}
%{\cal G\/}_{\Lambda_k}({\cal I\/}) &=& \sum_{i=1}^n p(k|x_i; \theta^b) \psi^{b^{-1}}\left[ \Lambda^b_k E\left[z_i z_i^{T}; \theta^b\right] - (x_i - \mu_k^b) E\left[z_i; \theta^b\right]^{T}\right] 
%\end{eqnarray}
%Substituting~(\ref{eq:mfa_e2}) and~(\ref{eq:mfa_e3}) in~(\ref{eq:mfa_mean})~(\ref{eq:mfa_lambda}) we see that the 
%scores simplify to,
\begin{eqnarray}
\label{eq:mfa_mean}
  {\cal G\/}_{\mu_k}({\cal I\/}) &=& \sum\nolimits_i h_{ik}
  \{S_k^b\}^{-1} \left(x_i - \mu^b_k\right) \\
  {\cal G\/}_{\Lambda_k}({\cal I\/}) &=& \sum\nolimits_i h_{ik}
  \Big[\{S_k^b\}^{-1}(x_i - \mu_k^b)(x_i - \mu^b_k)^T \beta_k^{b^{T}} \nonumber\\
  &-& \{S_k^b\}^{-1}\Lambda_k^b \Big]   \label{eq:mfa_lambda}
\end{eqnarray}
For a detailed discussion of the Q function, the reader is referred to the 
EM derivation in~\cite{ghahramani97}. Note that the scores with
respect to the means are functionally similar to the first order residuals
of~(\ref{eq:grad_mean}). However, the scores with respect to the factor
loading matrices $\Lambda_k$ account for covariance statistics of the
observations $x_i$, not just variances. We refer to~(\ref{eq:mfa_mean})
and~(\ref{eq:mfa_lambda}) as the MFA Fisher scores (MFA-FS).
%We refer to the vector
%obtained by concatenating~(\ref{eq:mfa_mean}) and~(\ref{eq:mfa_lambda})
%for all $k$ as the MFA Fisher score (MFA-FS). 
%Note that this is
%not a FV due to the absence of normalization by the
%Fisher information, which is more complex to compute than for the variance-GMM.
% Since, the computation of a Fisher information matrix, in order to convert second order gradients in~(\ref{eq:mfa}) is bit complicated, we do not convert MFA based representations into Fisher vectors and use them as score vectors. This is done quite often with gradient encodings in the literature~\cite{liu14}. 
% g

\subsubsection{MFA Fisher Information} 

The MFA-FV is obtained by scaling the MFA-FS by the Fisher information matrix. As before, this is approximated by a block-diagonal matrix that scales the Fisher scores of the $k^{th}$ mixture component by the inverse
square-root of 
\begin{equation}
\label{mfa_FIM}
{\cal F\/}_k = w_k Cov_k\left({\cal G\/}_k(x)\right).
\end{equation} 
Here $w_k$ is the weight of the $k^{th}$ mixture, ${\cal G\/}_k(x)$ the 
data term of its Fisher score, and $Cov_k$ the covariance with respect
to the $k^{th}$ mixture component. For the mean 
scores of~(\ref{eq:mfa_mean}) 
this is simply the component covariance $S^b_k$. For the factor loading 
scores it is the covariance of the data term of~(\ref{eq:mfa_lambda}).
This is a $D \times R$ matrix,  whose entry $(i, j)$ is the product of two
Gaussian random variables
\begin{eqnarray*}
{\cal G\/}^{(i, j)}_k(x) &=& f_i g_j \nonumber\\
&=& {\lfloor \{S_k^b\}^{-1} (x - \mu^b_k)\rfloor}_i
    {\lfloor\beta^b_k(x - \mu^b_k)\rfloor}_j 
\end{eqnarray*}
where ${\lfloor w \rfloor}_i$ is the $i^{th}$ element of vector 
$w$. The covariance matrix of the vectorized Fisher score is then
\begin{displaymath}
  Cov_k({\cal G\/}_k(x))_{(i, j),(l,m)} = 
  E\left[f_i g_j f_l g_m\right] - E[f_i g_j]E[f_l g_m].
\end{displaymath}
This can be simplified by using Isserlis' theorem,
%Isserlis, L. (1916). "On Certain Probable Errors and Correlation Coefficients of Multiple Frequency Distributions with Skew Regression". Biometrika. 11: 185–190. JSTOR 2331846. doi:10.1093/biomet/11.3.185.
which states that, for zero-mean Gaussian random variables 
$\{x_1, x_2, x_3, x_4\}$, $E[x_1 x_2 x_3 x_4] = E[x_1 x_2]E[x_3 x_4] + 
E[x_1 x_3]E[x_2 x_4] + E[x_1 x_4]E[x_2 x_3]$. It follows that
\begin{eqnarray}
\label{mfa_FScov}
  \lefteqn{Cov_k({\cal G\/}_k(x))_{(i, j),(l,m)} =} & & \\
  &=& E\left[f_i g_m\right]E\left[f_l g_j\right] + 
  E\left[f_i f_l\right]E\left[g_j g_m\right] \nonumber
\end{eqnarray}
with
\begin{eqnarray}
\label{mfa_Fexp}
  E\left[f_i g_m\right]E\left[f_l g_j\right] &=& 
  {\lfloor(S^b_k)^{-1}\Lambda^b_k\rfloor}_{i, m}
  {\lfloor(S^b_k)^{-1}\Lambda^b_k\rfloor}_{l, j}\nonumber\\
  E\left[f_i f_l\right]E\left[g_j g_m\right] &=& 
  {\lfloor(S^b_k)^{-1}\rfloor}_{i, l}
  {\lfloor\beta^b_k\Lambda^b_k\rfloor}_{j, m} 
\end{eqnarray}
The Fisher scaling of the $k^{th}$ MFA component is obtained by 
combining~(\ref{mfa_FIM}), (\ref{mfa_FScov}) and~(\ref{mfa_Fexp}). 

 \begin{figure}[t]
\begin{center}
  \includegraphics[width=\linewidth]{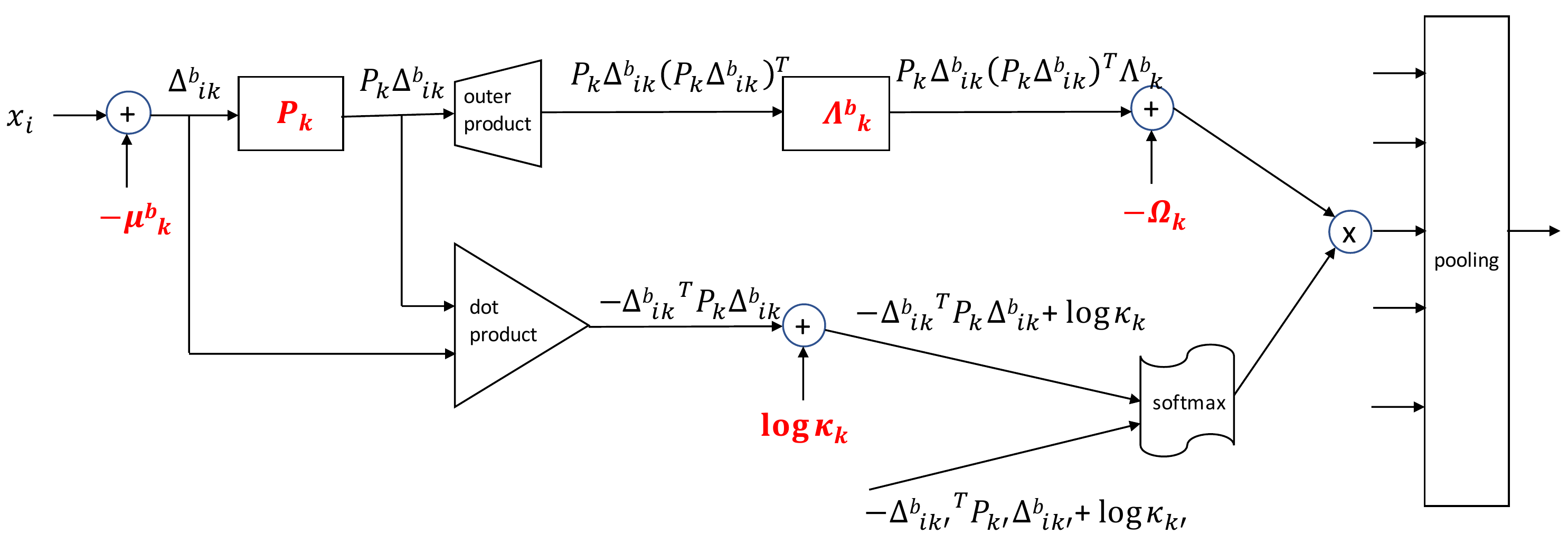}
\end{center}
\caption{\footnotesize{The MFA-FS($\Lambda$) layer implements
    \eqref{eq:mfaNet_lambda2} as a network layer. The bottom branch
    computes the posterior probability of \eqref{eq:mfaNet_prob2}. The top
    branch computes the remainder of the summation argument.
    Note that circles denote entry-wise operations, boxes implement
    matrix multiplications (weight layers), the outer product
    layer is similar to \cite{lin15}, and the dot-product layer a combination
    of elementwise multiplication and a sum. Expressions in red are
    parameters of the $k^{th}$ MFA component, in black the computations made
    by the network.}}
\label{fig:NetBD}
\end{figure}

Note that, like the GMM-FV, the MFA-FV can be applied with or
without the embedding into natural parameter space. However,
because it is still a Gaussian model, all arguments above suggest that
it should be more effective when applied after the embeddings of
(\ref{eq:nu1})-(\ref{eq:nu3}).

\section{Neural Network Embedding}
\label{sec:MFAFSNet}

The FVs above are implemented
independently of the CNN used to extract the SMNs.
The mixture model is learned from the extracted
BoS and the FV derived from its parameters. In this section,
we redesign the MFA-FS embedding as a CNN layer, to enable 
end-to-end training. 

\subsection{The MFA-FS Layer}

To implement \eqref{eq:mfa_mean} and \eqref{eq:mfa_lambda} in a CNN, 
we start by defining
\begin{equation}
\Delta_{ik}^b = x_i - \mu_k^b.
\label{eq:mfaNet_zeroed}
\end{equation}
%\begin{equation}
%S_k^b=\Lambda_k^b\Lambda_k^{b^T} + \psi^b
%\label{eq:mfaNet_cov}
%\end{equation}
Combining this with \eqref{eq:mfaNet_cov} and \eqref{eq:beta_k},
\eqref{eq:mfa_mean} and \eqref{eq:mfa_lambda} can be written as
\begin{align}
\label{eq:mfaNet_mean}
\mathcal{G}_{\mu_k}(\mathcal{I})&=\sum_ip(k|x_i; \theta^b)\{S_k^b\}^{-1}
\Delta_{ik}^b \\
\mathcal{G}_{\Lambda_k}(\mathcal{I})&=-\sum_ip(k|x_i; \theta^b) \{S_k^b\}^{-1}
   [\Delta_{ik}^b\Delta_{ik}^{b^T}\{S_k^b\}^{-1}\Lambda_k^b-\Lambda_k^b] \nonumber\\
   &=-\sum_i p(k|x_i; \theta^b) \{S_k^b\}^{-1}\Delta_{ik}^b [\{S_k^b\}^{-1}\Delta_{ik}^b]^T\Lambda_k^b
  \nonumber\\
  & \quad +\sum_i p(k|x_i; \theta^b)\{S_k^b\}^{-1}\Lambda_k^b 
  \label{eq:mfaNet_lambda}
\end{align}
Since the $k_{th}$ mixture component $p(x|s=k)$ has distribution
$G(x, \mu_k, S_k^b)$, it follows that
\begin{eqnarray}
\label{eq:mfaNet_prob}
p(k|x_i;\theta^b)&=&\frac{w_k G(x_i;\mu_k^b, S_k^b)}
  {\sum_kw_k G(x_i;\mu_k^b, S_k^b)} \\
  &=&\frac{\frac{w_k}{|S_k^b|^{\frac{1}{2}}}
    \exp\{-\frac{1}{2}\Delta_{ik}^{b^T}S_k^{b^{-1}}\Delta_{ik}^b\}}
  {\sum_k \frac{w_k}{|S_k^b|^{\frac{1}{2}}}
    \exp\{-\frac{1}{2}\Delta_{ik}^{b^T}S_k^{b^{-1}}\Delta_{ik}^b\}} \nonumber
\end{eqnarray}
and denoting
\begin{eqnarray}
  P_k&=&S_k^{b^{-1}}, \label{eq:mfaNet_Pk} \\
  \Omega_k&=&S_k^{b^{-1}}\Lambda_k^b, \label{eq:mfaNet_Omegak} \\
  \kappa_{k}&=&\frac{w_k}{|S_k^b|^{\frac{1}{2}}}, \label{eq:mfaNet_kappak}
\end{eqnarray}
finally leads to 
\begin{align}
\label{eq:mfaNet_mean2}
\mathcal{G}_{\mu_k}(\mathcal{I})&=\sum_i p(k|x_i; \theta^b)P_k\Delta_{ik}^b \\
\label{eq:mfaNet_lambda2}
  \mathcal{G}_{\Lambda_k}(\mathcal{I})
 &=-\sum_i p(k|x_i; \theta^b) 
  \{P_k\Delta_{ik}^b(P_k\Delta_{ik}^b)^T \Lambda_k^b-\Omega_k\} \\
 \label{eq:mfaNet_prob2}
\lefteqn{p(k|x_i;\theta^b) =\frac{\kappa_{k}\exp\{-\frac{1}{2}
                      \Delta_{ik}^{b^T}P_k\Delta_{ik}^b\}}
                      {\sum_{k'}\kappa_{k'}
                      \exp\{-\frac{1}{2}\Delta_{ik'}^{b^T}P_{k'}\Delta_{ik'}^b\}}} &&
\end{align}
Figure \ref{fig:NetBD} shows how \eqref{eq:mfaNet_lambda2} can be
implemented as a network layer. The bottom branch computes the
posterior probability of \eqref{eq:mfaNet_prob2}. The top branch computes
the remainder of the summation argument.
The computation of \eqref{eq:mfaNet_mean2} is similar. The bottom branch
is identical, the top branch omits the operations beyond
$P_k\Delta_{ik}^b$. However, because the benefits of this component are 
small, we only use the layer of Figure \ref{fig:NetBD}.

 \begin{figure}[t]\RawFloats
   \includegraphics[width=\linewidth]{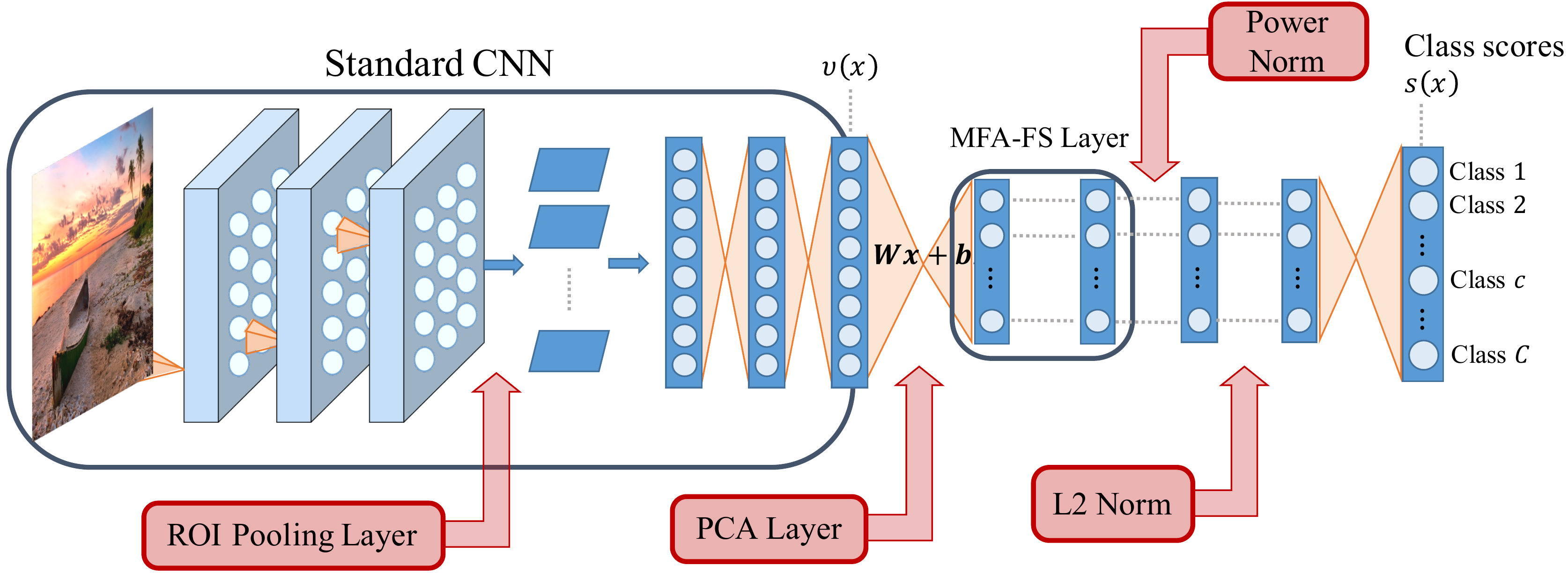}
   \caption{MFAFSNet architecture. A standard CNN, pretrained on ImageNet,
     is used to extract a vector $\nu(x)$ of image features. The network is
     applied to image patches, which are combined with a RoI pooling
     layer. The vector $\nu(x)$ is fed to a dimensionality reduction layer,
     then to the MFA-FS layer, and finally power and $L_2$ normalized,
     before application of a linear classification layer.}
   \label{fig:MFAMSNet}
 \end{figure}

\subsection{Network Architecture}

The overall architecture of the MFAFSNet is shown in Figure
\ref{fig:MFAMSNet}. A model pretrained on ImageNet is used to
extract a vector $\nu(x)$ of image features. This network is applied to 
image patches, producing multiple feature maps per image to classify. 
When the patches are of
a single scale, the model is converted to a fully convolutional network.
For patches of multiple scales, the final pooling layer is
replaced with a region-of-interest (ROI) pooling layer, which accepts
feature maps of multiple sizes and produces a fixed size output. This is a 
standard practice in object
detection~\cite{girshick14,girshick15}. The feature vector $\nu(x)$ is 
dimensionality reduced by a fc layer of appropriate dimensions,
and fed to the MFA-FS layer of Figure~\ref{fig:NetBD}.
Note that this layer pools multiple local features, corresponding to
objects of different sizes and in different image locations,
generating a single feature vector for the whole image.
This is fed to a power and a L2 normalization layers, and finally
to a linear classifier layer.

\subsection{Loss Function}

While the parameters $\mu_k^b$, $P_k$, $\lambda_k^b$, $\Omega_k$
and $\log \kappa_k$ are learned by back-propagation, they
must maintain their interpretation
as statistical quantities. This requires that \eqref{eq:mfaNet_cov} and
\eqref{eq:mfaNet_Pk}-\eqref{eq:mfaNet_kappak} hold. Some of these constraints
do not need to be enforced. For example, since \eqref{eq:mfaNet_kappak}
is the only to involve $w_k$, there is a one to one relationship
between $\log \kappa_k$ and $w_k$, independently of the value of
$|S_k^b|^{\frac{1}{2}}$. In result, it is equivalent to learn $w_k$ under the
constraint of \eqref{eq:mfaNet_kappak} or simply learn $\log \kappa_k$,
which leads to a simpler optimization. A similar observation holds
for \eqref{eq:mfaNet_cov}, which is the only constraint on $\psi^b$.
On the other hand, some of the relationships must be enforced to maintain the
MFA-FS interpretation. These are \eqref{eq:mfaNet_Pk},
\eqref{eq:mfaNet_Omegak}, and the symmetry of matrix $P_k$.
They are enforced by adding regularization terms to the loss function. 
For training set
${\cal D} = \{(x_i, y_i)\}$ and classification loss
$L_c(.)$, this leads to a loss function
\begin{eqnarray}
  L({\cal D}) &=&  L_c({\cal D}) +
  \lambda_1 \sum_k ||\Omega_k-P_k\Lambda_k^b||_F^2  \nonumber \\
  &+& \lambda_2 \sum_k ||P_k-P_k^T||_F^2
  \label{eq:mfaNet_loss}
\end{eqnarray}
where $||A||_F$ is the Frobenius norm of $A$, and $\lambda_1,\lambda_2$ 
control the  regularization strength. We use the hinge loss
\begin{equation}
  L_c({\cal D}) = \frac{1}{N}\sum_{i=1}^N\sum_{k=1}^K
  \left[\max(0, 1-I(y_i,k)s_k(x_{i})\right]^2
  \label{eq:hinge}
\end{equation}
where $s(x)$ is the input to the softmax at the top of the network
and $I(\cdot)$ the indicator function. This is for consistency with the 
FV literature, which is based on SVMs. Any other classification loss could 
be used.

\section{Experimental Results}

In this section, we present the results of an extensive evaluation of all
the FVs discussed above.

\subsection{Experimental set-up}
\label{sec:setup}

All experiments are based on the MIT Indoor~\cite{dset:MITIndoor} and 
SUN~\cite{dset:MITSUN} scene datasets. MIT Indoor consists of 100 images 
each from 67 indoor scene classes. Following the standard protocol, we use 
80 images per class for training and the remaining 20 for testing. MIT SUN 
has about 100K images from 397 indoor and outdoor scene categories. It 
provides randomly sampled image sets each with 50 images per class for 
training as well as test. Performance, on both datasets, is reported as 
average per class classification accuracy. 

For the FVs implemented independently of the object recognition CNN, the 
BoS is extracted with  the CNNs of~\cite{krizhevsky12} 
or~\cite{simonyan2014very}, pre-trained on ImageNet. The networks are 
applied convolutionally, generating a 1,000 dimensional SMN for roughly 
every 128x128 pixel region. The 1,000 dimensional probability vectors are 
reduced to 500 dimensions using PCA. 
These are the descriptors $\pi_i$ used to build the FV. Unless otherwise 
noted, GMM-FVs, DMM-FVs, and logSMN-FVs use a $100$-component mixture model.
All FVs are power and $L_2$ normalized~\cite{perronnin10}, and classified with
a linear SVM. Unless otherwise noted, the MFA is computed with $K=50$
components and latent space dimension $R=10$.

The MFAFSNet is implemented with the object recognition networks of
\cite{krizhevsky12,simonyan2014very,he15}, trained on ImageNet. The 
vector $\nu(x)$ of Figure~\ref{fig:MFAMSNet} is the input to the softmax at 
the top of these networks, i.e. we use the $\nu^{(2)}$ embedding 
of~\eqref{eq:nu2}. A vector $\nu(x)$  is produced per $l\times l$ image 
patch, mapped into $500$ dimensions by the PCA layer, and fed to the
MFA-FS layer. Images are resized, reducing 
smaller side to $512$-pixels and maintaining aspect ratio.
Three patch sizes, $l\in\{96, 128, 160\}$ were used, producing between
$590$ and $1000$ patches per image. The MFA-FS layer uses $K=50$ 
mixture components and $R=10$ subspace dimensions, outputting a vector of
$500\times50\times10$ dimensions. The last fc layer is learned with a
learning rate of $0.001$, while $0.00001$ is used for all others.
Momentum and weight decay were set to $0.9$ and $0.0005$ respectively and
the network trained on 10 epochs. 
For simplicity, we set $\lambda_1=\lambda_2=\lambda$ in 
\eqref{eq:mfaNet_loss}.

\begin{figure}[t]\RawFloats
  \begin{minipage}{0.56\linewidth}
    \captionof{table}{Comparison of FV encodings. 
      SIFT-FV is a GMM-FV applied to
      a bag of SIFT features, and conv5-FV a GMM-FV applied at
      layer conv5.}
    \label{tab:smnFVs}
    \centering
    \small
    \begin{tabular}{|c|c|c|}
      \hline
      Method & MIT & SUN \\
      & Indoor & \\
      \hline
      SIFT-FV & 60.0 & 43.3\\
      conv5-FV & 61.4 &  -  \\
      SMN-FV & 55.3 &  36.87 \\
      DMM-FV  & 58.8 &   40.86 \\
      logSMN-FV & \textbf{67.7} & \textbf{49.86} \\
      \hline
    \end{tabular}
  \end{minipage}
  \begin{minipage}{.43\linewidth}
    \captionof{table}{Ablation analysis on MIT Indoor, for the
      role of centroids: learned, transferred (T), or random (R).
    %Transfer means that GMM parameters are
    %learned with a DMM and converted analytically to the GMM format,
    %and vice-versa.
  }
  \label{tab:codebooks}
  \centering
  \small
  \begin{tabular}{|c|c|}
    \hline
    FV & Accuracy  \\
    \hline
    DMM &  58.8 \\
    DMM(T) &  58.4 \\
    DMM(R) &  58.0 \\
    logSMN &  68.5\\
    logSMN(T) &  \textbf{68.6}\\
    logSMN(R) & 61.3\\
    \hline
  \end{tabular}
\end{minipage}
\end{figure}

\subsection{Benefits of Natural Parameter Space}
\label{sec:piFV}

We begin by comparing FVs that embed the BoS distribution, the GMM-FV
(denoted SMN-FV) and DMM-FV, to the logSMN-FV, which embeds natural 
parameter descriptors. For completeness, we also tested 
the classical SIFT-FV of~\cite{sanchez13} and a GMM-FV that embeds the 
features of CNN convolutional layer 5, denoted conv5-FV. 
The CNN is that of~\cite{krizhevsky12}. All GMM-FVs use a reference 
GMM $\theta^b$ and are computed with~(\ref{eq:fv_mean}). 
We ignore the variance component of~(\ref{eq:fv_variance}), since
it did not improve the performance. The DMM-FV uses a reference 
DMM $\theta^b = \{\alpha_k^b, w_k^b\}_{k=1}^K$  
and is computed with (\ref{eq:grad_alpha}) and~(\ref{eq:FIM_dmm}). 
The logSMN-FV is computed with~\eqref{eq:fvlog_mean}.

Table~\ref{tab:smnFVs} summarizes the performance of all methods. 
The SMN-FV is a poor classifier, 
underperforming the SIFT and conv5 FVs by~$5-6\%$ points. This is surprising, 
given the now well documented advantage of CNN over SIFT features and the 
increased discriminant power of SMN over conv5 features.
The DMM-FV outperforms the SMN-FV but still underperforms the other two
approaches. The limited improvement 
over the SMN-FV can be partially explained by
%fig~\ref{fig:DMM_FV},
%which shows a plot of the classification performance as a function
%of the number of mixture components
the somewhat surprising observation that best performance is achieved with 
a single component mixture. This is unlike GMM, which tends to 
improve significantly with the number of components, and suggests that the 
DMM lacks the flexibility to fit complex distributions. 
On the other hand, the logSMN-FV leads to a staggering
improvement in classification accuracy, beating the
SMN-FV by around $9\%$ on both datasets!
This is surprising, since it is identical to the
SMN-FV, up to the log mapping into the natural parameter space.
It is also the first BoS FV to outperform the SIFT-FV.

\subsection{Ablation Analysis}  
\label{exp:ablation}

To better understand these differences, we did an ablation 
analysis of the parameters of Table \ref{tab:relations} on MIT Indoor.

{\bf Centroids:} Consider a DMM
$\{w_k, \alpha_k\}_{k=1}^K$ learned from SMNs $\pi_i$ and a GMM
$\{w_k, \mu_k, \Sigma_k\}_{k=1}^K$ learned from logSMNs $\log \pi_i$.
Since, in this case, \eqref{eq:muk} and \eqref{eq:fk} are identical, the
GMM centroids can be used to construct a DMM and vice versa. For the
former, it suffices to map the Gaussian means $\{\mu_k\}_{k=1}^K$ in
$\log \pi$ space to the Dirichlet parameters, by solving 
$\mu_k = \psi\left(\tilde{\alpha}_k\right) -
\psi\left(\sum_k \tilde{\alpha}_k\right)$ for $\tilde{\alpha}_k$. The
GMM $\{w_k, \mu_k, \Sigma_k\}_{k=1}^K$ can then be mapped to a
DMM $\{w_k, \tilde{\alpha}_k\}_{k=1}^K$, using the estimated
$\tilde{\alpha}_k$ and copying weights $w_k$.
For the latter, a GMM is anchored at the DMM centroids
$\tilde{\mu}_k = f_k(\alpha)$, the weights $w_k$
copied from the latter, and the Gaussian covariances $\Sigma_k$
set to the global covariance $\Sigma$ of the training data. We refer
to this process as model transfer.

We trained a DMM $\{w_k, \alpha_k\}_{k=1}^K$ and a GMM
$\{w_k, \mu_k, \Sigma\}_{k=1}^K$ in $\log \pi$ space.
A second Dirichlet model, DMM(T), of parameters
$\{w_k, \tilde{\alpha}_k\}_{k=1}^K$ was obtained by transferring the
GMM means.  Similarly, a second Gauss mixture, GMM(T),
of parameters $\{w_k, \tilde{\mu}_k, \Sigma\}_{k=1}^K$ was transferred
from the DMM. These models were compared to a DMM of random centroids, 
DMM(R),  and a GMM of random centroids in $\log \pi$ space 
and covariances $\Sigma_k$ set to the global covariance $\Sigma$ of the
training data, GMM(R). The mixture weights of DMM(R) and GMM(R) 
were set to uniform.

\begin{table}\RawFloats
%\vspace{0.1mm}
  \captionof{table}{Ablation analysis on MIT Indoor for the role of scaling
    and assignments. Tilde indicates transferred parameters.}
\label{tab:ablation}
\centering
\small
\begin{tabular}{|c|c|c|c|}
\hline
Mixture & \multicolumn{3}{c|}{FV Encoding} \\
\cline{2-4}
Model & Scaling & Assignment & Accuracy \\
\hline
\hline
\multirow{4}{*}{DMM} & \multirow{2}{*}{${\cal F}_k^{-1/2}(\alpha)$} & 
   $q_k(\pi; \alpha, w)$ & 58.8 \\
\cline{3-4}
 & & $h_k(\log\pi; \tilde{\mu}, \Sigma, w)$ &  57.7\\
\cline{2-4}
 & \multirow{2}{*}{$\frac{1}{\sqrt{w_k} \sigma_k}$} 
     & $q_k(\pi; \alpha, w)$ & 67.1 \\
\cline{3-4}
 & & $h_k(\log\pi; \tilde{\mu}, \Sigma, w)$ &  68.6\\
\hline
\hline
\multirow{4}{*}{GMM} & \multirow{2}{*}{$\frac{1}{\sqrt{w_k} \sigma_k}$} & 
     $h_k(\log\pi; \mu, \Sigma, w)$ &  68.5\\
\cline{3-4}
 &  & $q(\pi; \tilde{\alpha}, w)$ & 68.7 \\
\cline{2-4}
 & \multirow{2}{*}{${\cal F}_k^{-1/2}(\tilde{\alpha})$} & 
  $h_k(\log\pi; \mu, \Sigma, w)$ & 57.7 \\
\cline{3-4}
 &  & $q(\pi; \tilde{\alpha}, w)$ & 58.4 \\
\hline
\end{tabular}
\end{table}

Table~\ref{tab:codebooks} summarizes the performance of all FVs.
The DMM-FV is always substantially weaker than the logSMN-FV.
Nevertheless, the logSMN(T) result shows that the DMM can be used to
estimate the mixture parameters. In fact, the GMM with centroids
transferred from the DMM has slightly better performance than the original GMM.
On the other hand, learning the logSMN GMM and transferring to the DMM-FV
form has weak performance. These results show that the
{\it accuracy of the estimation of the mixture model is much less important
  than the FV encoding itself.\/} This is further confirmed by the results
of the models with random parameters. For the weaker DMM-FV, random
centroids perform as well as original or transferred centroids. In fact, all
these FVs underperform the stronger logSMN-FV encoding with no
learning (random centroids).
%In summary, no approach to learning the mixture overcomes the limitations of a
%weak FV encoding. 

In summary, it is more important to use a stronger FV encoding (logSMN-FV)
than carefully learn parameters of a weaker FV encoding (DMM-FV).
However, for the stronger logSMN-FV, there is a non-trivial gain 
in using learned centroids. On the other hand, it does not matter if they 
are the GMM centroids or transfered from a DMM.
All of this follows from the equality of \eqref{eq:muk} and \eqref{eq:fk} for
the logSMN-FV and the DMM-FV. Since this equality also holds for the 
residuals $\nu(\pi_i) - \xi(\theta_k)$
of~\eqref{eq:FV_general}, the two FV encodings only differ in
the scaling $\gamma(\theta_k)$ and assignments ($h_k(\log \pi; \mu, \Sigma, w)$
vs $q_k(\pi; \alpha, w)$).

{\bf Scaling and assignments:}
The impact of the two factors can be
studied by starting from the logSMN-FV and changing the scaling function
to ${\cal F}_k^{-1/2}$ or the assignment function to 
$q_k(\pi; \tilde{\alpha}, w)$.
The combination of the two modifications leads to the DMM(T)-FV. 
Conversely, it is possible to start from the DMM-FV and change
each of the functions or both, in which case we obtain the logSMN(T)-FV.
Table~\ref{tab:ablation} shows that the assignment function has
a small effect. In all cases, the gain of changing assignment 
is at most 1.5\%, and no assignment
function is clearly better. What seems to matter is that it {\it matches the
scaling function.\/} The DMM performs better with $q_k(.)$ for the original
${\cal F}_k^{-1/2}$ scaling, but with $h_k(.)$ for Gaussian scaling. The GMM 
has equivalent performance with the two assignments for
Gaussian scaling, but performs better with $q_k(.)$ for ${\cal F}_k^{-1/2}$
scaling.

On the other hand, {\it the scaling function has a significant effect
on classification accuracy.\/} For both mixtures and assignment functions,
Gaussian scaling is 10\% better than ${\cal F}_k^{-1/2}$ scaling.
These results are not totally surprising, since Gaussian scaling has
an additional degree of freedom (variance $\sigma_k$), while
${\cal F}_k^{-1/2}$ is determined by the $\alpha$ parameters
already used to determine centroids. The normalization
of Gaussian scaling, which  produces normal residuals of zero mean and unit 
variance, is akin to batch normalization~\cite{ioffe2015batch}, which is 
well known to benefit learning.
It is also known that the most important effect of Fisher information 
scaling is the decorrelation of the FV, which improves its performance 
significantly~\cite{sanchez13}. On the other hand, (\ref{eq:FIM_dmm})
shows that the scaling matrix ${\cal F}_k$ has a restrictive structure, 
in that its off-diagonal elements are equal to $-\psi'(\sum_l \alpha_{kl})$. 
Hence, ${\cal F}_k$ resides in a 
subspace of the space of symmetric positive definite matrices 
$\mathbb{S}_d^{+}$ and affords very few degrees of freedom 
(roughly equal to the dimensionality of $\alpha$).

Since scaling is determined by the Fisher information and this defines
the local metric on the tangent space ${\cal T\/}_{\theta^b}$ to the model
manifold, these results suggest that the Dirichlet manifold is not
suited for classification. It is, however, interesting that
classification is so strongly affected by the choice of model
manifold. This is particularly surprising because, as shown in
Table~\ref{tab:smnFVs}, Gaussian scaling is not a top performer 
in the absence of the $\log \pi$
embedding. In summary, while effective classification requires logSMNs, 
the modeling manifold is well captured by the GMM.

\begin{table}\RawFloats
%\vspace{0.1mm}
\caption{Accuracy of GMM-FVs implemented with the natural parameter
embeddings of~(\ref{eq:nu1})-(\ref{eq:nu3}).}
\label{tab:nus}
\centering
\small
\begin{tabular}{|c|c|c|}
\hline
NP embedding & MIT Indoor & SUN \\
\hline
$\nu^{(1)}$ & 67.7 & 50.87 \\
$\nu^{(2)}$ &  \textbf{68.5} &  \textbf{51.17} \\
$\nu^{(3)}$ &  67.6 &  50.47\\
\hline
\end{tabular}
%\vspace{-0.3mm}
\end{table}

\begin{figure}[t]
\centering
  \includegraphics[scale=0.35]{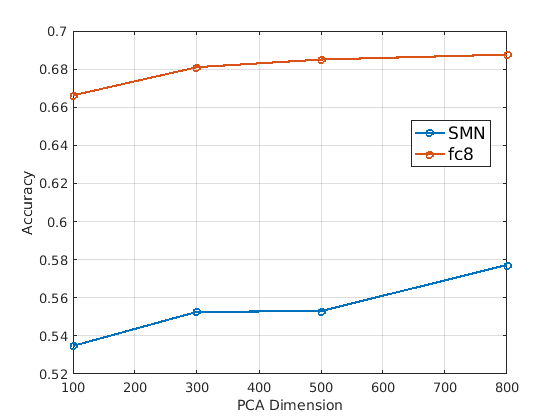}
\caption{Performance variation of the $\nu^{(2)}$-FV and the SMN-FV with PCA dimension.
}
\label{fig:PCAvar}
\end{figure}

\subsection{Natural Parameter Embedding}
\label{exp:npFV}

The vastly superior performance of the logSMN-FV in Table~\ref{tab:smnFVs}
suggests that the FV should be computed in natural parameter space.
In fact, it corresponds to the natural parameter transformation
of~\eqref{eq:nu1}. We next compared the transformations 
of~(\ref{eq:nu1})-(\ref{eq:nu3}).
Table~\ref{tab:nus} summarizes the performance of the FV implemented
with each mapping. $\nu^{(2)}$ has the best performance, followed by the 
$\nu^{(1)}$ (the logSMN-FV) and $\nu^{(3)}$. This is consistent with 
the fact that the softmax CNN is trained to maximize discrimination in the 
space of $\nu^{(2)}$, as discussed in Section~\ref{sec:npBoS},
% In result, the $\nu^{(2)}$-FV is recommended for semantic features generated by softmax CNNs. These
confirming that 1) probability modeling is difficult
in the simplex, and 2) natural parameter transformations alleviate the problem. The performance of the $\nu^{(2)}$ embedding does not vary drastically with the dimensionality of the PCA transformation used before FV encoding. As shown in fig~\ref{fig:PCAvar}, even with 100 dimensional PCA projection, the $\nu^{(2)}$-FV achieves an accuracy of $~66\%$, which is much higher than the SMN-FV.  

Finally, to ensure that the gains of the FV are not just due to the use of a log non-linearity, we applied the log transformation to the activations of the penultimate CNN layer~(often referred to as fc7) and $\nu^{(2)}$ features in~(\ref{eq:nu2}), that already reside in the NP space. Rather than a gain, this resulted in a substantial decrease in performance ($58\%$ with an log fc7-FV vs $65.1\%$ with an fc7-FV and $60.97\%$ with a $\log \nu^{(2)}$-FV vs $68.5\%$ with a $\nu^{(2)}$-FV, on MIT Indoor scenes). This was expected, since the role of log is natural parameter transformation and the argument does not apply for spaces other than the probability simplex.

%Table~\ref{tab:nus} summarizes the accuracy of
%GMM-FVs derived with all the NP embeddings of~(\ref{eq:nu1})-(\ref{eq:nu3}), denoted as $\nu^{(i)}$-FVs. As expected, the $\nu^{(2)}$-FV has 
%best performance, followed by the logSMN-FV ($\nu^{(1)}$) and 
%the $\nu^{(3)}$-FV. The differences are within 1\%, which is
%is consistent with the known benefits of end-to-end
%training. In result, the $\nu^{(2)}$-FV is recommended for semantic
%features generated by softmax CNNs.

%It's vastly superior performance to the SMN-FV, therefore, 

\begin{figure*}[t]\RawFloats
  \begin{minipage}{.28 \linewidth}
    \captionof{table}{Scene classification accuracy of various embeddings.}
    \label{tab:GMM_v_MFA}
    \centering
    \small
    \begin{tabular}{|c|c|c|}
      \hline
      Descriptor & MIT & SUN \\
      & Indoor & \\
      \hline
      \hline
      \multicolumn{3}{|c|}{Object-based} \\
      \hline
      \hline      
      GMM FV ($\mu$) & 66.08 & 50.01\\
      \hline
      GMM FV ($\sigma$) & 53.86 &  37.71\\
      \hline
      MFA FS ($\mu$) & 67.68 &  51.43\\
      \hline
      MFA FS ($\Lambda$)  & \textbf{71.11} & \textbf{53.38}  \\
      \hline
      MFA FV ($\mu$) & 66.73 & 51.37 \\
      \hline
      MFA FV ($\Lambda$)  & 70.89 & 53.56 \\
      \hline 
      \hline
      \multicolumn{3}{|c|}{Gist-based} \\
      \hline
      \hline      
      BoS-fc1  & 64.84 & 47.47\\
      \hline
      BoS-fc2  & 69.36 &  50.9\\
      \hline
      BoS-fc3 & 70.6 &  53.12\\
      \hline
    \end{tabular}
  \end{minipage}
  \begin{minipage}{.34 \linewidth}
    \centering
    \includegraphics[width = \linewidth]{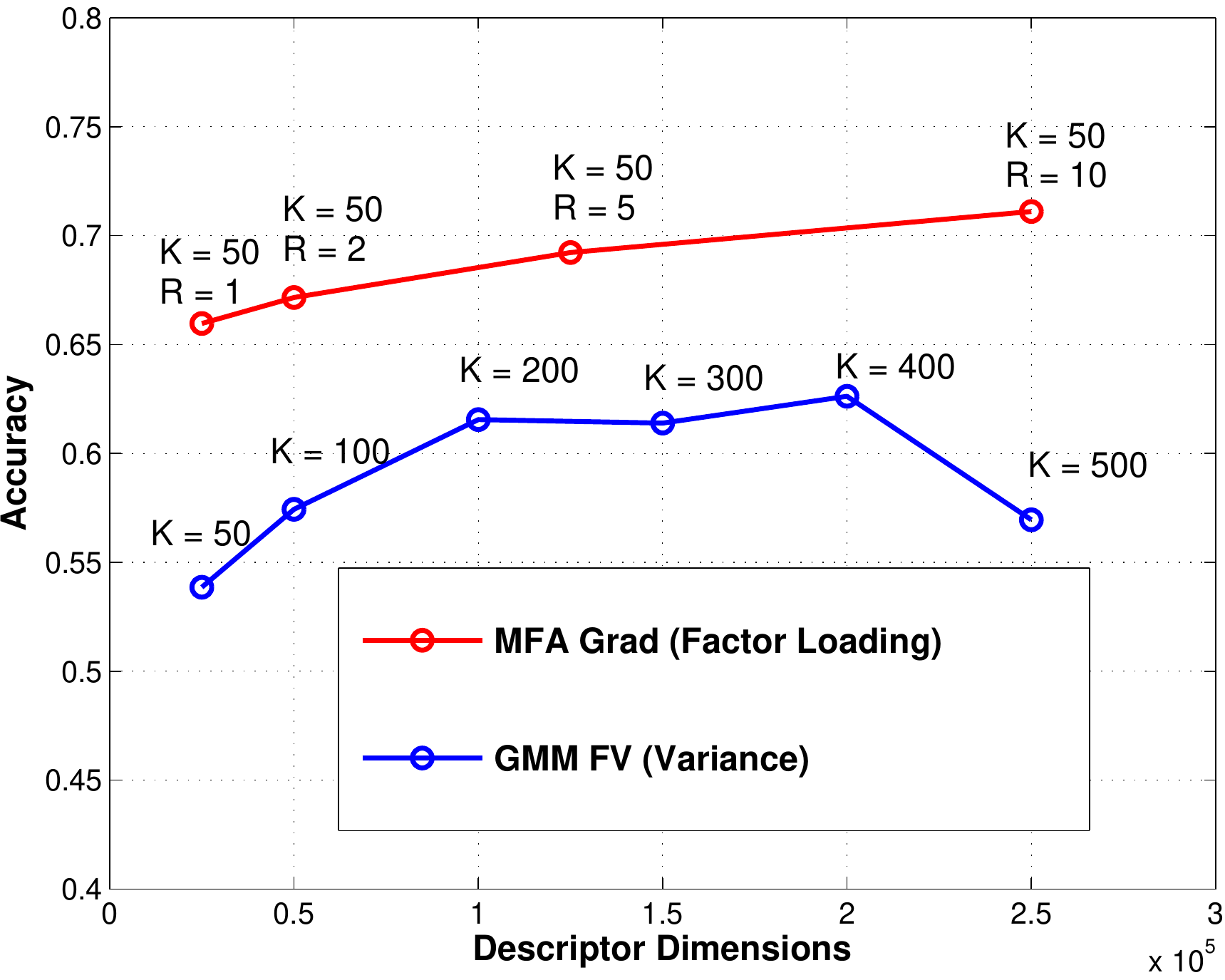} 
    \caption{Accuracy vs. descriptor size for MFA-FS($\Lambda$) 
      of $50$ components and $R$ factor dimensions and 
      GMM-FV($\sigma$) of $K$ components.}
    \label{fig:Cov_vs_var}
  \end{minipage}
  \begin{minipage}{0.34\linewidth}
    \centering
    \includegraphics[width=\linewidth]{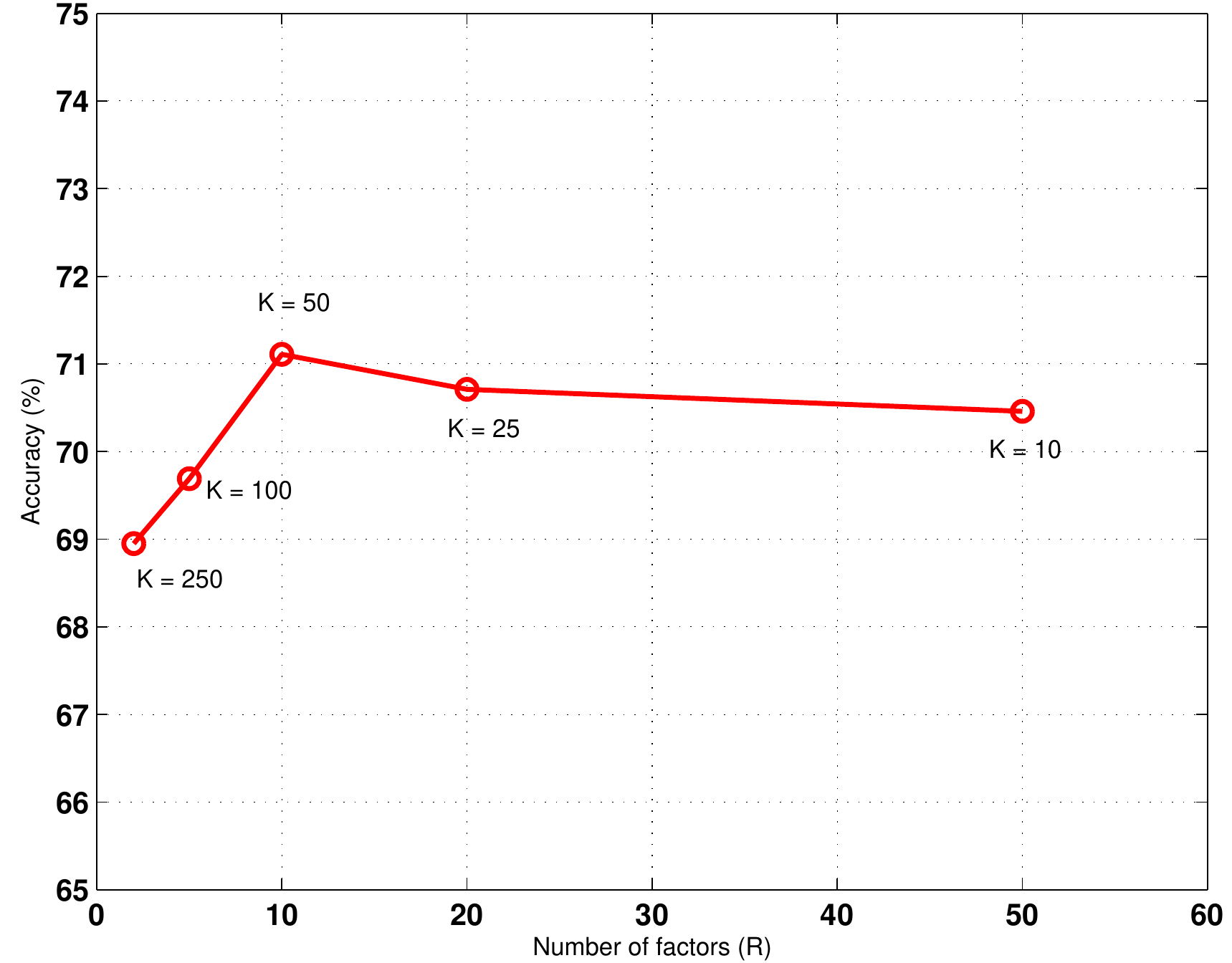}
    \caption{Accuracy of MFA-FSs of constant size $K \times R$ (components
      vs. factors). From left to right, $R$ incraeses while $K$ 
      decreases.}
    \label{fig:Local_v_Global}
  \end{minipage}
\end{figure*}

\subsection{Comparison to previous embeddings}

Other embeddings have been proposed for the classification of semantic vectors. \cite{kwitt12} projects SMNs on the great circle, using a square root embedding $\sqrt{\pi}$. The non-Euclidean nature of the simplex and the non-linearity of its geodesics are noted as a major difficulty for SMN 
based classification. The square root embedding was also used 
in~\cite{delhumeau13} for SIFT descriptors. Rather than  $L_2$ normalization, the authors propose $L_1$ normalization of the SIFT histogram to produce a probability vector. The SIFT probabilities are then transformed into
``Root-SIFT'' descriptors and encoded with a GMM-FV. This achieved moderate improvements over standard SIFT. We applied the square root embedding to the SMNs, achieving $58.95\%$ accuracy on MIT Indoor and $40.6\%$ on SUN. This is close to the DMM-FV results of Table \ref{tab:smnFVs}, but drastically inferior to the $\nu^{(2)}$-FV results of Table~\ref{tab:nus}. The inability of the root embedding to replicate the linearization of Figure~\ref{fig:natural} d) limits the performance of the GMM-FV after the embedding.

An alternative embedding of probability descriptors was 
introduced in~\cite{kobayashi14}. It uses a $\log$ 
transformation on $L_1$ normalized SIFT descriptors and is also inspired by Dirichlet sufficient statistics. A SIFT probability vector $p$ is subjected to a von-Mises transformation 
$\nu^{(5)} = \frac{\log(p + \epsilon) - \log\epsilon}{\|\log(p + \epsilon) 
  - \log\epsilon\|_2}$ and encoded by a GMM-FV. This was shown to improve
on the rootSIFT-FV of~\cite{delhumeau13}. Except for its $L_2$ 
normalization, the von-Mises embedding is somewhat similar to
natural parameter transformation $\nu^{(3)} = \log p_i - \log p_N$. When we
applied it to SMNs, it achieved $63.4\%$ on MIT Indoor and $46.1\%$ on
SUN. This was better than the square root embedding, but underperformed
all three embeddings of~(\ref{eq:nu1})-(\ref{eq:nu3}). The most likely reason 
is the projection onto the great circle ($L_2$ normalization), which may work for SIFT but does not help for CNN semantics.

%{\bf While 
%this derivation of the MFA information has a tutorial value, in practice it did not result in any gains over simply using the scores of $\Lambda$ as our image representation. Therefore, we avoid scaling the scores in~ in~(\ref{eq:mfa_lambda}) for our image classification experiments, although they may be useful in some other scenarios.} [THIS NEEDS TO BE SHOWN WITH RESULTS.]

\subsection{Covariance Modeling}

We next evaluate the importance of covariance modeling, by studying
the MFA-FV derived from a MFA learned in the natural parameter space.
Unless otherwise noted, results refer to MIT Indoor.

\subsubsection{Importance of Covariance Modeling}

The MFA was compared to the variance-GMM, using the set-up of 
Section~\ref{sec:setup}, embedding $\nu^{(2)}$ of \eqref{eq:nu2},
$K=50$ components and latent space dimension $R = 10$. 
Table~\ref{tab:GMM_v_MFA} compares the GMM-FV($\mu$)
of~\eqref{eq:fv_mean}, the GMM-FV($\sigma$) of~\eqref{eq:fv_variance}, the
MFA-FS($\mu$) of ~\eqref{eq:mfa_mean}, the 
MFA-FS($\Lambda$) of~\eqref{eq:mfa_lambda}, the MFA-FV($\mu$) which
scales the MFA-FS($\mu$) with $(w_k S^b_k)^{-1/2}$ and the MFA-FV($\Lambda$) 
which scales the MFA-FS($\Lambda$) with the Fisher information
of (\ref{mfa_FIM})-(\ref{mfa_Fexp}). 
GMM-FV($\sigma$) was the weakest performer, underperforming
the GMM-FV($\mu$) by more than $10\%$. This difference is much larger
for CNN features than for the lower dimensional SIFT
features~\cite{perronnin10} and the reason why CNN FVs only consider
gradients w.r.t. means~\cite{gong14,dixit15}. 

The improved covariance modeling of the MFA solves this 
problem. The MFA-FS($\Lambda$) significantly outperforms both 
GMM-FVs and the MFA-FS($\mu$). 
% Note that, due to covariance modeling in MFAs, the MFA-FS($\mu$) 
% outperforms the GMM-FV($\mu$). In the MFA-FS($\mu$) pooling 
% of~(\ref{eq:mfa_mean}), the first order residuals are scaled by 
% covariance matrices rather than scalar variances. This local de-correlation 
% explains the non-trivial improvement of the MFA-FS($\mu$) over the 
% GMM-FV($\mu$)($\sim 1.5\%$ points). 
A related covariance modeling was used in~\cite{tanaka13} to 
obtain FVs w.r.t. Gaussian means and local subspace variances~(covariance
eigenvalues). In our experiments, this subspace variance FV 
($60.7\%$ on MIT Indoor) outperformed the variance GMM-FV($\sigma$) but
was clearly inferior to the MFA-FS($\Lambda$), which captures full 
covariance. In summary, full covariance modeling appears to be essential for
FV-style pooling of CNN features.

On the other hand, the MFA-FV($\mu$) and MFA-FV($\Lambda$) have similar performance. Unlike the GMM-FV, where variance gradients
are uninformative but Fisher scaling has large gains, the MFA-FV derives
most of its power from the covariance gradients (MFA-FS($\Lambda$)) and
gains little from Fisher scaling. In fact, even the concatenation
(MFA-FS($\mu$), MFA-FS($\Lambda$)) gave small
improvement~($\sim 1\%$), which does not justify the increased computation.
We use the MFA-FS($\Lambda$) alone in the following sections.

\subsubsection{Covariance Modeling vs Subspace Dimensions}

The comparison above is somewhat unfair because, for fixed number of
components $K$, the GMM-FV has less parameters than the MFA-FS.
Fig.~\ref{fig:Cov_vs_var} compares the GMM-FV($\sigma$) and MFA-FS($\Lambda$)
when $K$ varies in $\{50, \ldots, 500\}$ and the MFA latent space
dimensions $R$ in $\{1, \ldots, 10\}$. For comparable dimensions, the
covariance based scores significantly outperform the variance statistics.
%While these results are for MIT Indoor we observed the same behavior on SUN.
% the experiments were repeated for a mixture of $K=500$ components,
% the accuracy of the GMM-FV ($\mu$) classifier increased to $69.9\%$ on 
% MIT Indoor and $-\%$ on SUN. This, however, was still inferior to
% the performance of the MFA ($\Lambda$) of $K=50$ components. 
% {\bf The variance FV's of GMMs, as shown in , however, did not improve with increasing mixture sizes. }
%A final observation is that the MFA-FS($\mu$) has similar performance to 
%the GMM-FV($\mu$). This is not surprising, since these gradients are 
%functionally similar. 
Fixed-size MFA-FS~($\Lambda$) descriptors (250K dimensions)
were next used to evaluate the relative importance of local covariance
modeling (dimensionality $R$) and global flexibility of the
mixture (components $K$). MFA models were learned with $K$ decreasing from 
250 to 10, each reduction in $K$ being traded for an increase in $R$, from 2 
to 50. As shown in Figure~\ref{fig:Local_v_Global}, classification accuracy
increased steadily as $K$ decreased from 250 to 50 ($R$ increasing
from 2 to 10). This shows that there is a trade-off between local
covariance modeling and global manifold approximation, as suggested by
Figure~\ref{fig:MFA_example}. 
With better covariance modeling, fewer mixture components are required to
cover the manifold of semantic features. Obviously, if $R$ is too large
the number of components may not be enough to enable a global 
approximation of the manifold. 
It appears, however, that low subspace dimensionality is more costly
than few components. For small $R$, even models 
with large $K$ ($K=250$) perform poorly. Best results are achieved 
when the MFA has a sufficient number of Gaussians that implement a 
reasonable linear approximation of the local manifold. In our experiments,
this corresponded to $K = 50, R = 10$.

\subsubsection{Gist Descriptors}

The MFA-FS follows a tradition of embeddings that pool a 
bag-of-descriptors~\cite{nuno00,csurka04,lazebnik06,perronnin10}. 
The underlying i.i.d. assumptions make the embedding 
flexible, with no template-like rigidity. An alternative
scene representation is a holistic ``gist'' descriptor, e.g., 
produced by a fully connected (fc) neural network layer. The MFA-FS
was compared to gist embeddings based on a network of one or more fc 
layers, interspersed with layers of ReLu non-linearities.
The BoS was used to produce a tensor of 10 x 10 x 1000 responses, containing
1000 dimensional $\nu^{(2)}$ descriptors extracted from roughly every 
128x128 image region. An fc layer was then used to map
the tensor into a 4096 dimensional vector and followed by a ReLU layer. 
Successive fc layers of 4096 input and 4096 output channels and ReLU
stages were optionally added to create a deeper embedding. The final 4096 
dimensional vector was fed to a linear classifier, trained with a scene 
classification loss. 
All fc layers were learned with ``drop-out'' of probability 0.2. 
The embeddings are denoted BoS-fc1 to  BoS-fc3 based on the number 
of fc layers used. 

%\begin{table}[t]
%\vspace{0.1mm}
% \caption{Comparison of MFA-FS with semantic ``gist'' embeddings learned 
%   using ImageNet BoS and the Places dataset.}
% \label{tab:Places_embedding}
% \centering
% \begin{tabular}{c|c|c}
% \hline
% \hline
% Descriptor & MIT Indoor & SUN \\
% \hline
% MFA FS ($\Lambda$)  & \textbf{71.11} & \textbf{53.38}  \\
% \hline
% \hline
% BoS-fc1  & 64.84 & 47.47\\
% %\hline
% BoS-fc2  & 69.36 &  50.9\\
% %\hline
% BoS-fc3 & 70.6 &  53.12\\
% \hline
% \hline 
% \end{tabular}
% %\vspace{0.2mm}
% \end{table}

A problem for this approach is the limited amount of
transfer data. The number of parameters in the largest embedding 
was almost equal to that of the ImageNet CNN
of~\cite{krizhevsky12, simonyan14}. MIT Indoor is too small to 
train such an embedding, leading to a scene classification accuracy of
$33\%$. To overcome the problem, we used the Places scene 
dataset~\cite{zhou14}, which contains 2.4 M training images
of $~200$ scene categories. This, however, made the training of
gist embeddings last several days on a GPU,
as opposed to two hours for the MFA-FS. Nevertheless, as shown
in Table~\ref{tab:GMM_v_MFA}, the BoS-fc embeddings 
%were used as image representation for transfer based 
%scene classification, on MIT Indoor and SUN. 
%shows that, despite their complexity, they 
still underperformed the MFA-FS.

\subsection{MFAFSNet}

We finish with a set of experiments on the MFAFSNet.

\subsubsection{Relevance of statistical interpretation}

\begin{figure}[t]\RawFloats
  \begin{minipage}{.45\linewidth}
    \centering 
    \captionof{table}{Effect of initialization on classification accuracy.}
    \small
    \setlength{\tabcolsep}{4pt}
    \begin{tabular}{|c|c|c|}
      \hline
      & MIT & SUN \\
      &Indoor& \\
      \hline
      \hline
      & \multicolumn{2}{c|}{AlexNet} \\
      \hline
      \hline
      Random& 69.82 & 50.23 \\
      MFA-FS & 71.44 &54.14 \\
      \hline
      \hline
      & \multicolumn{2}{c|}{VGG-16}\\
      \hline \hline
      Random& 77.3 & 56.2\\
      MFA-FS & 80.3 & 62.51 \\
      \hline
    \end{tabular}
    \label{tab:mfaNet_init}
  \end{minipage}
  \begin{minipage}{.54\linewidth}
    \includegraphics[width =\linewidth]{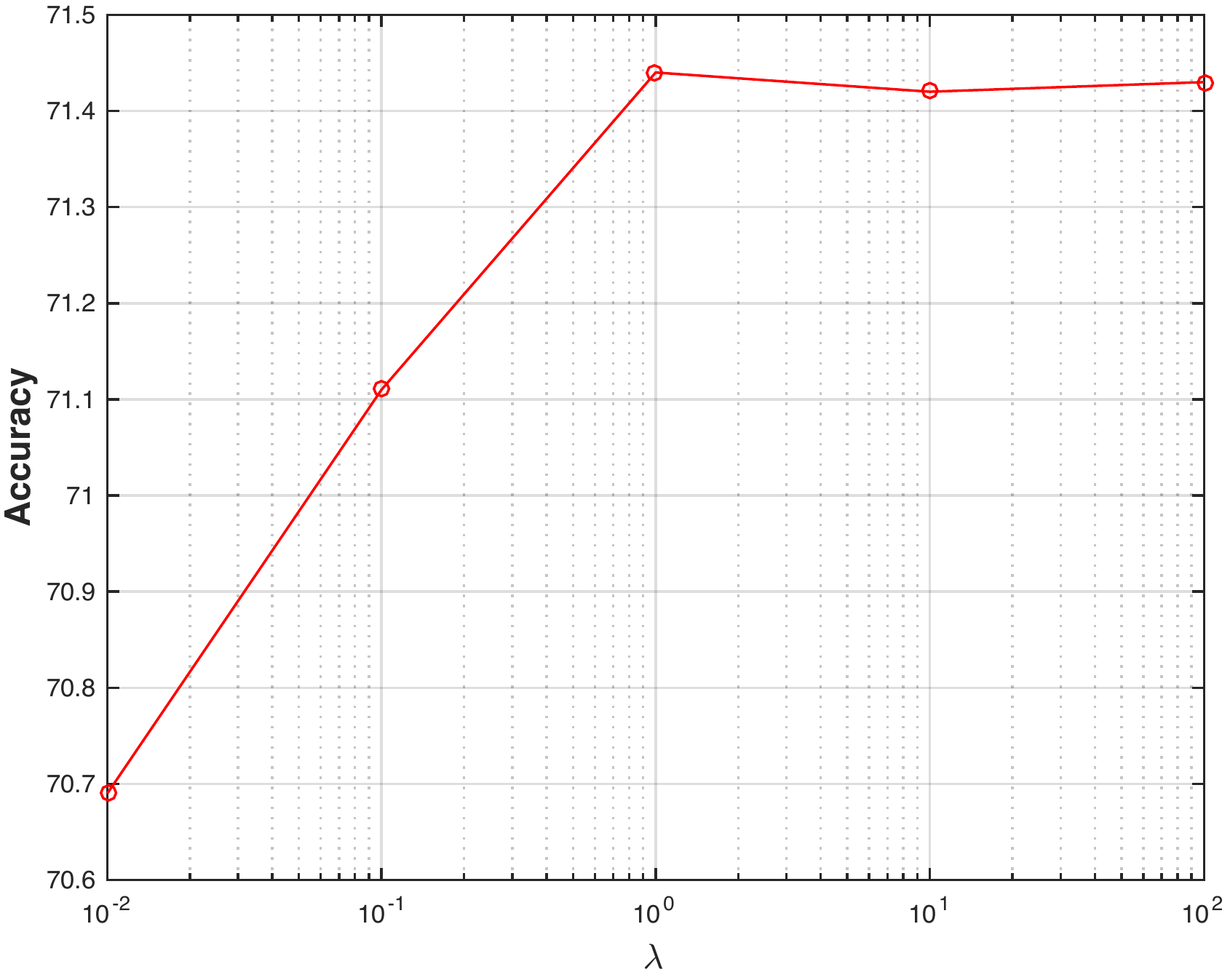} 
    \caption{Effect of regularization on classification accuracy.}
    \label{fig:reg}
  \end{minipage}
\end{figure}

A set of experiments was conducted to test the need to enforce the statistical 
interpretation of the MFAFSNet\footnote{Results are reported for a 
single patch size of $96\time96$, but similar behavior was observed for 
other configurations.}.
The first addressed parameter initialization, comparing
random initialization of the MFA-FS parameters (zero mean
Gaussian of standard deviation $0.01$) to initialization
with the MFA-FS (PCA matrix learned 
from all patches $\nu(x)$ and MFA layer learned by 
EM~\cite{ghahramani97}).  A strength $\lambda = 1$ was used 
in \eqref{eq:mfaNet_loss}.  Table \ref{tab:mfaNet_init} shows that random
initialization was weaker by $2-3\%$ on MIT Indoor and $4-6\%$ 
on SUN.  The importance of regularization was next investigated
 by varying $\lambda$.  For small $\lambda$, the learning algorithm is
free to ignore the MFA-FS constraints. For larger $\lambda$, the
network has stronger statistical interpretation. Figure \ref{fig:reg}
shows an improvement of up to $1\%$ when $\lambda$
increases from $0.01$ to $1$.
% \begin{table}[t]
%   \caption{Effect of regularization strength on MFAFSNet
%     classification accuracy.}
%     \label{tab:reg}
%   \centering
%   \begin{tabular}{|c|c|c|c|c|c|}
%     \hline
%     \multicolumn{6}{|c|}{AlexNet}\\
%     \hline
%     \hline
%     $\lambda$&0.01&0.1&1&10&100\\
%     \hline
%     Accuracy& 70.69 & 71.11 &71.44& 71.42& 71.43\\
%     \hline
%     \hline
%     \hline
%     \multicolumn{6}{|c|}{VGG-16}\\
%     \hline
%     \hline
%     $\lambda$& 0.01 &0.1 &1 &10 &100\\
%     \hline
%     Accuracy & 79.79& 80.19& 80.3& 80.12&80.14\\
%     \hline
%   \end{tabular}
% \end{table}
These experiments show that it is important to enforce the statistical
interpretation of the MFAFSNet. In all remaining experiments we use
MFA-FS initialization and $\lambda = 1$.

\subsubsection{Multi-scale and end-to-end learning}

A set of experiments then investigated the impact of multiple patch sizes.
%, so as to account for the fact that objects can 
%occur at multiple scales in scenes.
Table~\ref{tab:MFA+CNNs} compares the accuracies of the 
MFA-FS($\Lambda$) and MFAFSNet with 96x96, 128x128 and 160x160 
patches, as well as their combination (3 scales). 
For the MFA-FS, the three vectors were concatenated. For the MFAFSNet,
a mixture of bounding boxes of the three sizes was fed to the ROI pooling 
layer. This generated a shorter vector, better suited for
the available GPU memory. The
multi-scale combination achieved the best performance for all CNNs 
and datasets. This is not surprising, as it accounts for multiple
object sizes within the scenes. The consistently better performance 
of the MFAFSNet, over the MFA-FS, also confirms the benefits of 
end-to-end learning.

\subsubsection{Comparison to previous transfer-based methods}

Various methods have been proposed to transfer ImageNet object classifiers
to scenes~\cite{liu14,cimpoi15a,dixit15,gao16}. Since they
only report results for MIT Indoor, we compare results for this 
dataset only on Table~\ref{tab:MIT_v_Objects}. 
%The GMM-FV of~\cite{dixit15} operates on AlexNet CNN semantics extracted from image patches of multiple sizes~(96, 128, 160, 80). 
The GMM FV of~\cite{cimpoi15a} uses convolutional features 
from AlexNet or VGG-16 extracted in a large multi-scale setting.
% Their best results using a single scale are $68.5\%$ on MIT Indoor and $50.03\%$ on SUN. Using a combination of 4 patch scales this improves to $72.86\%$ and $54.4\%$ respectively. Our MFA based classifier outperforms them with accuracies of $71.1\%$ and $53.54\%$ (on MIT Indoor and SUN resp.) for a single patch scale. For a 3 scale combination our method achieves $73.58\%$ on MIT Indoor and $55.95\%$ on SUN, which again exceeds their performance.
\cite{liu14} proposed a gradient representation based on sparse codes
and reported results for a single patch scale of 128x128 
and AlexNet features. 
%In this setting, our MFA based representation performs better than their classifier on the MIT Indoor scenes ($71.1\%$ v $68.2\%$). 
An improved H-Sparse representation, 
combining multiple scales and VGG features was later proposed in~\cite{liu16}.
%, achieves $79.5\%$ accuracy. Our VGG based multi-scale MFA still outperforms their proposal with an $81.43\%$ accuracy. 
The recent bilinear (BN) pooling method of~\cite{lin15} is similar to the 
MFA-FS in that it captures global second order descriptor statistics. 
%Their results reproduced in~\cite{gao16} using VGG-16 network features, are $77.55\%$. 
The simplicity of these descriptors enables fine-tuning
of CNN layers to scene classification. However, as shown in~\cite{gao16} for 
VGG-16 features, the results 
are clearly inferior to those of the MFA-FS without fine-tuning and 
about $5\%$ worse than the MFAFSNet.
%Yet our MFA based representation that cannot avail of fine-tuning outperforms their method with a non-trivial margin~(single scale acc:$79.57\%$, multi-scale acc:$80.1\%$). 
\cite{gao16} proposes to compress these bilinear 
statistics with trainable transformations. However, the resulting image 
representation of size $8K$ has accuracy inferior to
combining the MFA-FS with a PCA of $5K$ dimensions. 
In summary, the MFA-FS and MFAFSNet are state of the
art procedures for task transfer from object recognition (on ImageNet) 
to scene classification (on MIT Indoor/SUN). 
%As seen above,
%the end-to-end training of the MFAFSNet improves the results of MFA-FS 
%by $1-2\%$. {\bf even through the MFA-FS concatenates Fisher vectors of 
%different patch scales, to form a vector that is much longer than that 
%of the MFAFSNet.} 
The closest competitor \cite{cimpoi15a} combines CNN features in a 
massive multiscale setting (10 image sizes). The MFA-FS and MFAFSNet 
outperform it with only 3 scales.

\begin{figure}\RawFloats
  \centering
  \captionof{table}{Classification accuracy as a function of 
      patch size $p \times p$. 'All' denotes combination of
      three sizes.}
    \label{tab:MFA+CNNs}
  \setlength{\tabcolsep}{2pt}
  \small
  \begin{tabular}{|c|c||c|c|c|c||c|c|c|c|}
    \hline
    &&\multicolumn{4}{c||}{MIT Indoor} & \multicolumn{4}{c|}{SUN} \\
    \hline
    && 160 & 128 & 96 & All     & 160 & 128 & 96 & All\\
    \hline
    \hline
    MFA-FS & \multirow{2}{*}{AlexNet} & 69.8 & 71.1 & 70.5 & 73.6 
                      & 52.4 & 53.4 & 53.5 & 56.0 \\
    MFAFSNet & & 70.1& 71.9 & 71.5 & {\bf 75.3} 
                      & 52.6 & 54.5 & 54.2 & {\bf 57.3}\\
    \hline
    MFA-FS & \multirow{2}{*}{VGG-16} & 77.3 & 77.3 & 80.0 & 80.1 & 59.8 
                                & 61.0 & 61.7 & 63.3 \\
    MFAFSNet &  & 78.3& 78.8 & 80.5 & {\bf 81.3} 
                      & 61.5 & 62.0 & 61.7 & {\bf 64.8}\\
    \hline
    MFA-FS & \multirow{2}{*}{ResNet-50} & 81.2 & 82.0 & 82.4 & 83.4 
                      & 63.5 & 65.2 & 64.8 & 65.7\\
    MFAFSNet & & 81.5 & 82.7 & 83.0 & {\bf 84.0} 
                      & 63.4& 65.5 & 65.7 & {\bf 66.3}\\
    \hline
  \end{tabular}
\end{figure}

\subsubsection{Task vs. dataset transfer}
\label{sec:task_transfer}

The MFA based classifiers implement task transfer, using
an object recognition network to classify scenes. This is an alternative 
to the standard dataset transfer, where a network trained to classify scenes
is applied to a different scene dataset. This approach is simpler but much more
intensive, requiring the collection and annotation of a large scene
dataset. It was pursued in~\cite{zhou14}, which assembled
a scene dataset (Places) of 2.4M images and used it to train scene 
classification CNNs. These were then transfered to MIT Indoor and SUN, by 
using the the CNN as a features extractor and linearly
classifying these features.

Table~\ref{tab:Imnet_v_Places} compares the performance of the two 
transfer approaches. Somewhat surprisingly, task transfer with the 
MFAFSNet {\it outperformed\/} dataset transfer with the pre-trained Places CNN, on both 
datasets, for all networks. We include an 
additional baseline, denoted Places ft, where in addition to large scale training on scenes, the CNN is finetuned to Indoor and SUN as well. The transfer based MFAFSNet 
beats this strong baseline for AlexNet and the deeper Resnet-50 architectures, and is only slightly worse for the VGG architecture. 
% This supports the hypothesis 
% that the variability of configurations of most scenes makes scene classification more challenging than object recognition, to the point where CNN architectures of strong object recognition performance are much less effective for scenes. It is, instead, preferable to pool object detections across the scene image as done by the MFAFSNet.
An ensuing question is whether there is any complementarity between the object-based MFAFSNet and the holistic representation learned by the Places 
CNN. This was tested by training a classifier on the concatenation of the two descriptors. As shown in Table~\ref{tab:Imnet_v_Places}, it lead to a 
substantial increase in performance ($6-8\%$), suggesting that the representations are indeed {\it very complementary.\/} To the best of our 
knowledge, no method using these
or deeper CNNs has reported better results on these datasets.
This is detailed in Table~\ref{tab:SOA}, which compares results to recent scene classification methods in the literature. The MFAFSNet + Places combination is a state-of-the-art classifier with substantial gains over all other approaches. 

\begin{figure}[t]\RawFloats
  \begin{minipage}{.55\linewidth}
  \centering
  \captionof{table}{Task transfer performance on MIT Indoor. 'All' denotes
  combined sizes $96, 128, 160$.}
  \label{tab:MIT_v_Objects}
  \footnotesize
    \setlength{\tabcolsep}{2pt}
      \begin{tabular}{|c|c|c|c|c|}
        \hline
        Method& \multicolumn{2}{c|}{Scales} & \multicolumn{2}{c|}{Scales} \\
              & $128$ & All & $128$ & All  \\
        \hline\hline
        &\multicolumn{2}{|c|}{AlexNet}&\multicolumn{2}{|c|}{VGG} \\
        \hline
        MFAFSNet & {\bf 71.85} & {\bf 75.31} & {\bf 80.48} & {\bf 81.32}\\
        \hline
        MFA-FS & 71.11 & 73.58 & 79.9 & 81.43\\
%        \hline
%        GMM FV \cite{dixit15}& 68.5 & 72.86\\
        \hline
         FV+FC \cite{cimpoi15a} & - & 71.6 & - & 81.0 \\
%        ~\cite{dixit15} & & \\
        \hline
        % FV (Alexnet)(3 scales)~\cite{dixit15} & 71.24\\
        Sp. Code & 68.2 & - & - & 77.6 \\
        \cite{liu14,liu16} &  &  &  &  \\
        \hline        
        H-Sparse & -&- &- & 79.5 \\
        \cite{liu16}& &  & &  \\
        \hline
        BN \cite{gao16} & -& - & 77.55 & - \\
        %\hline
        %FV+FC \cite{cimpoi15a} &-& -& - & 81.0 \\
        \hline\hline
        & \multicolumn{4}{|c|}{VGG + dim. reduct.} \\
        \hline\hline
        MFA-FS + & - & - & {\bf 79.3} & -\\
        PCA (5k) & & &  & \\
        \hline
        BN (8k) &- &- & 76.17 & - \\
        \cite{gao16} & & & & \\
        \hline
      \end{tabular}
  \end{minipage}
  \begin{minipage}{.43\linewidth}
    \centering
    \captionof{table}{Task vs. dataset transfer.
      'Both' referes to the combination of MFAFSNet 
      with Places CNN.}
      \footnotesize
    \label{tab:Imnet_v_Places}
    \setlength{\tabcolsep}{2pt}
    \begin{tabular}{|c|c|c|}
      \hline
      & SUN & Indoor \\
      \hline
      \hline
      &\multicolumn{2}{c|}{AlexNet} \\
      \hline
      \hline
      Places & 54.3 & 68.24\\
      \hline
      %MFA-FS  & 55.95 & 73.58\\
      %Both & 63.16 & 79.86  \\
      Places ft & 56.8 & 72.16 \\
      \hline
      MFAFSNet  & 57.29 & 75.31 \\
      Both& \textbf{64.47} & \textbf{80.49} \\
      \hline
      \hline
      &\multicolumn{2}{|c|}{VGG} \\
      \hline
      \hline
      Places & 61.32 & 79.47\\
      \hline
      %MFA-FS & 63.31 & 80.1\\
      %Both & 71.06 & 87.23 \\
      Places ft & 65.25& 81.34\\
      \hline
      MFAFSNet & 64.81 & 81.32\\
      Both& \textbf{72.43} & \textbf{88.05} \\
      \hline
      \hline
      &\multicolumn{2}{|c|}{Resnet-50} \\
      \hline
      \hline
      Places & 63.51 & 79.05\\
      \hline
      Places ft & 63.80 & 82.61\\
      \hline
      MFAFSNet & 66.28 & 83.99\\
      Both& \textbf{73.35} & \textbf{88.06} \\
      \hline
    \end{tabular}
  \end{minipage}
\end{figure}

\section{Conclusion}

This work makes several contributions to computer vision. First, we
introduced a new task transfer architecture based on sophisticated
pooling operators for CNN features, implemented under the FV paradigm.
While good performance was demonstrated for object-to-scene transfer, 
the architecture is applicable to any problems involving the transfer of
a set of source semantics into a set of target semantics that are loose 
combinations of them. Image captioning~\cite{xu2015show} or visual 
question-answering~\cite{antol2015vqa} are examples of vision problems that 
could leverage such transfer.

Second, we demonstrated the importance of semantic representations for this 
type of transfer. While others had argued for this in the 
past~\cite{antol2015vqa}, the semantic noise of pre-CNN semantic 
spaces prevented the implementation of effective semantic transfer systems. 
We have shown that the combination of the BoS produced by a CNN and 
a sophisticated transfer architecture enable state of the art performance 
in problems like scene classification. In fact, this transfer was 
shown to {\it outperform\/} the direct learning of CNNs from much
larger scene datasets. 
%This shows that the
%solution of computer vision problems does not reduce to collecting large
%datasets and training CNNs. 
While transfer learning has been pursued as a vehicle for efficient 
training, the results above indicate that task transfer could be 
essential to the solution of complex vision problems. This points towards
modular vision systems and is an agreement with human cognition, which is
highly modular and rich in interactions of modules specialized in different
semantics.

Third, we have contributed evidence to the long standing debate on whether
scene understanding is based on objects or gist. Here, the most
significant finding was the amplitude of the gains of
combining the two representations. While this could be due to
sub-optimality of our object or gist-based solutions, it is unlikely
that better training or larger datasets would suffice to overcome
the large gap between the individual and joint performances. This makes
intuitive sense, since an object-representation of scenes must combine
localized detections in a manner invariant to object configurations. As 
we have shown, this requires very non-linear pooling operators that are 
complicated to learn. In the absence of explicit object supervision, 
a CNN could find it difficult to uncover them. On the other hand, a holistic
gist component appears to be critical as well. For example,
it accounts for the relative placement of objects in the
scene. 

\begin{figure}[t]\RawFloats
  \captionof{table}{Performance of scene classification methods.}
    \label{tab:SOA}
    \small
    \setlength{\tabcolsep}{2pt}
      \begin{tabular}{|c|c|c||c|c|c|}
        \hline
        & Indoor & SUN & & Indoor & SUN\\
        \hline \hline
        \multicolumn{3}{|c|}{AlexNet} &  \multicolumn{3}{|c|}{VGG} \\
        \hline        \hline
        \multicolumn{6}{|c|}{Without Places} \\
        \hline        \hline
        Sparse Cod.~\cite{liu14} & 68.2 & - &
        DAG-CNN ~\cite{yang15} &77.5 &56.2\\
        VLAD~\cite{gong14} &68.88 &51.98 &
        Sparse Cod.~\cite{liu16} & 77.6 & - \\
        Mid Level~\cite{li15}& 70.46& - &
        Compact BN~\cite{gao16} & 76.17 & - \\
        FV+FC ~\cite{cimpoi15a} &71.6 & -&
        Full BN~\cite{gao16} & 77.55 & - \\
        MFA-FS & 73.58 & 55.95 &
        H-Sparse~\cite{liu16} & 79.5 & - \\
        MFAFSNet & 75.01 & 57.15 &
        FV+FC~\cite{cimpoi15a} &81.0 & - \\
         & & & MFA-FS & 81.43 & 63.31 \\
        & & &  MFAFSNet & 82.66 & 64.59 \\
        \hline        \hline
        \multicolumn{6}{|c|}{With Places} \\
        \hline        \hline
        MetaClass~\cite{wu15} &78.9 &58.11 & Places+SF~\cite{khan2017scene}&84.3 &67.6 \\
         & & & LS-DHM~\cite{guo2017locally}&83.75 &67.56 \\
        MFA-FS& 79.86 & 63.16 & 
        MFA-FS & 87.23 & 71.06\\
        MFAFSNet& \textbf{80.49} & \textbf{64.47}& 
        MFAFSNet & \textbf{88.05} & \textbf{72.43}\\
        \hline
       % FV (AlexNet)(4 scales) + Places~\cite{dixit15} & 79.0 &61.72\\
       % FV (AlexNet)(3 scales) + Places~\cite{dixit15} &78.5$^*$ &-\\
       % FV (AlexNet) (4 scales)~\cite{dixit15} &72.86 & 54.4\\
       % FV (Alexnet)(3 scales)~\cite{dixit15} & 71.24 & 53.0\\
      \end{tabular}
\end{figure}

\ifCLASSOPTIONcompsoc
  \section*{Acknowledgments}
\else
  \section*{Acknowledgment}
\fi

This work was supported by NSF awards IIS-1208522 and IIS-1637941 and
GPU donations by Nvidia.

\ifCLASSOPTIONcaptionsoff
  \newpage
\fi



\bibliographystyle{IEEEtranS}

\bibliography{semanticsbib}

% Generated by IEEEtranS.bst, version: 1.14 (2015/08/26)
\begin{thebibliography}{10}
\providecommand{\url}[1]{#1}
\csname url@samestyle\endcsname
\providecommand{\newblock}{\relax}
\providecommand{\bibinfo}[2]{#2}
\providecommand{\BIBentrySTDinterwordspacing}{\spaceskip=0pt\relax}
\providecommand{\BIBentryALTinterwordstretchfactor}{4}
\providecommand{\BIBentryALTinterwordspacing}{\spaceskip=\fontdimen2\font plus
\BIBentryALTinterwordstretchfactor\fontdimen3\font minus
  \fontdimen4\font\relax}
\providecommand{\BIBforeignlanguage}[2]{{%
\expandafter\ifx\csname l@#1\endcsname\relax
\typeout{** WARNING: IEEEtranS.bst: No hyphenation pattern has been}%
\typeout{** loaded for the language `#1'. Using the pattern for}%
\typeout{** the default language instead.}%
\else
\language=\csname l@#1\endcsname
\fi
#2}}
\providecommand{\BIBdecl}{\relax}
\BIBdecl

\bibitem{adelson01}
\BIBentryALTinterwordspacing
E.~H. Adelson, ``On seeing stuff: the perception of materials by humans and
  machines,'' \emph{Proc. SPIE}, vol. 4299, pp. 1--12, 2001. [Online].
  Available: \url{http://dx.doi.org/10.1117/12.429489}
\BIBentrySTDinterwordspacing

\bibitem{akata16}
Z.~Akata, F.~Perronnin, Z.~Harchaoui, and C.~Schmid, ``Label-embedding for
  image classification,'' \emph{TPAMI}, vol.~38, no.~7, pp. 1425--1438, 2016.

\bibitem{antol2015vqa}
S.~Antol, A.~Agrawal, J.~Lu, M.~Mitchell, D.~Batra, C.~Lawrence~Zitnick, and
  D.~Parikh, ``Vqa: Visual question answering,'' in \emph{ICCV}, 2015, pp.
  2425--2433.

\bibitem{bergamo14}
A.~Bergamo and L.~Torresani, ``Classemes and other classifier-based features
  for efficient object categorization,'' \emph{TPAMI}, p.~1, 2014.

\bibitem{bergamo11}
A.~Bergamo, L.~Torresani, and A.~Fitzgibbon, ``Picodes: Learning a compact code
  for novel-category recognition,'' in \emph{NIPS}, J.~Shawe-Taylor, R.~Zemel,
  P.~Bartlett, F.~Pereira, and K.~Weinberger, Eds., 2011, pp. 2088--2096.

\bibitem{blei03}
\BIBentryALTinterwordspacing
D.~M. Blei, A.~Y. Ng, and M.~I. Jordan, ``{Latent Dirichlet Allocation},''
  \emph{Journal of Machine Learning Research}, vol.~3, pp. 993--1022, January
  2003. [Online]. Available:
  \url{http://www.jmlr.org/papers/volume3/blei03a/blei03a.pdf}
\BIBentrySTDinterwordspacing

\bibitem{cimpoi15a}
M.~Cimpoi, S.~Maji, I.~Kokkinos, and A.~Vedaldi, ``Deep filter banks for
  texture recognition, description, and segmentation,'' \emph{IJCV}, 2015.

\bibitem{cimpoi15}
M.~Cimpoi, S.~Maji, and A.~Vedaldi, ``Deep filter banks for texture recognition
  and segmentation,'' in \emph{CVPR}, June 2015.

\bibitem{cinbis12}
R.~Cinbis, J.~Verbeek, and C.~Schmid, ``Image categorization using fisher
  kernels of non-iid image models,'' in \emph{Computer Vision and Pattern
  Recognition (CVPR), 2012 IEEE Conference on}, 2012, pp. 2184--2191.

\bibitem{csurka04}
G.~Csurka, C.~R. Dance, L.~Fan, J.~Willamowski, and C.~Bray, ``Visual
  categorization with bags of keypoints,'' in \emph{In Workshop on Statistical
  Learning in Computer Vision, ECCV}, 2004, pp. 1--22.

\bibitem{delhumeau13}
J.~Delhumeau, P.~H. Gosselin, H.~J{\'e}gou, and P.~P{\'e}rez, ``Revisiting the
  vlad image representation,'' in \emph{ACM Multimedia}, 2013, pp. 653--656.

\bibitem{DLR}
A.~P. Dempster, N.~M. Laird, and D.~B. Rubin, ``Maximum likelihood from
  incomplete data via the em algorithm,'' \emph{Journal of the Royal
  Statistical Society, B, 39, 1-38}, 1977.

\bibitem{dset:ImNet}
J.~Deng, W.~Dong, R.~Socher, L.-J. Li, K.~Li, and L.~Fei-Fei, ``Imagenet: A
  large-scale hierarchical image database,'' in \emph{CVPR}, June 2009, pp.
  248--255.

\bibitem{dixit15}
M.~Dixit, S.~Chen, D.~Gao, N.~Rasiwasia, and N.~Vasconcelos, ``Scene
  classification with semantic fisher vectors,'' in \emph{CVPR}, June 2015.

\bibitem{feifei-lda05}
L.~Fei-Fei and P.~Perona, ``A bayesian hierarchical model for learning natural
  scene categories,'' in \emph{CVPR}, vol.~2, jun. 2005, pp. 524 -- 531 vol. 2.

\bibitem{gao16}
Y.~Gao, O.~Beijbom, N.~Zhang, and T.~Darrell, ``Compact bilinear pooling,''
  \emph{CoRR}, vol. abs/1511.06062, 2015.

\bibitem{maaten11}
L.~Getoor and T.~Scheffer, Eds., \emph{ICML}.\hskip 1em plus 0.5em minus
  0.4em\relax Omnipress, 2011.

\bibitem{ghahramani97}
Z.~Ghahramani and G.~E. Hinton, ``The em algorithm for mixtures of factor
  analyzers,'' Tech. Rep., 1997.

\bibitem{girshick15}
R.~Girshick, ``Fast r-cnn,'' in \emph{ICCV}, December 2015.

\bibitem{girshick14}
R.~Girshick, J.~Donahue, T.~Darrell, and J.~Malik, ``Rich feature hierarchies
  for accurate object detection and semantic segmentation,'' in \emph{CVPR},
  2014.

\bibitem{gong14}
Y.~Gong, L.~Wang, R.~Guo, and S.~Lazebnik, ``Multi-scale orderless pooling of
  deep convolutional activation features,'' in \emph{ECCV}, vol. 8695, 2014,
  pp. 392--407.

\bibitem{guo2017locally}
S.~Guo, W.~Huang, L.~Wang, and Y.~Qiao, ``Locally supervised deep hybrid model
  for scene recognition,'' \emph{TIP}, vol.~26, no.~2, pp. 808--820, 2017.

\bibitem{he_sppnet14}
K.~He, X.~Zhang, S.~Ren, and J.~Sun, ``Spatial pyramid pooling in deep
  convolutional networks for visual recognition,'' in \emph{ECCV}, D.~Fleet,
  T.~Pajdla, B.~Schiele, and T.~Tuytelaars, Eds.\hskip 1em plus 0.5em minus
  0.4em\relax Cham: Springer International Publishing, 2014, pp. 346--361.

\bibitem{he15}
------, ``Deep residual learning for image recognition,'' \emph{CoRR}, vol.
  abs/1512.03385, 2015.

\bibitem{ioffe2015batch}
S.~Ioffe and C.~Szegedy, ``Batch normalization: Accelerating deep network
  training by reducing internal covariate shift,'' in \emph{ICML}, 2015, pp.
  448--456.

\bibitem{jakkola99}
T.~S. Jaakkola and D.~Haussler, ``Exploiting generative models in
  discriminative classifiers,'' in \emph{NIPS}, 1999, pp. 487--493.

\bibitem{jain15}
M.~Jain, J.~C. van Gemert, and C.~G.~M. Snoek, ``What do 15,000 object
  categories tell us about classifying and localizing actions?'' in
  \emph{CVPR}, June 2015, pp. 46--55.

\bibitem{jain_o2a15}
M.~Jain, J.~C. van Gemert, T.~Mensink, and C.~G.~M. Snoek, ``Objects2action:
  Classifying and localizing actions without any video example,'' \emph{CoRR},
  vol. abs/1510.06939, 2015.

\bibitem{jegou10}
\BIBentryALTinterwordspacing
H.~J\'egou, M.~Douze, C.~Schmid, and P.~P\'erez, ``Aggregating local
  descriptors into a compact image representation,'' in \emph{CVPR}, jun 2010,
  pp. 3304--3311. [Online]. Available:
  \url{http://lear.inrialpes.fr/pubs/2010/JDSP10}
\BIBentrySTDinterwordspacing

\bibitem{khan2017scene}
S.~H. Khan, M.~Hayat, and F.~Porikli, ``Scene categorization with spectral
  features,'' in \emph{CVPR}, 2017, pp. 5638--5648.

\bibitem{kobayashi14}
T.~Kobayashi, ``Dirichlet-based histogram feature transform for image
  classification,'' in \emph{CVPR}, June 2014.

\bibitem{kordumova2016pooling}
S.~Kordumova, T.~Mensink, and C.~G. Snoek, ``Pooling objects for recognizing
  scenes without examples,'' in \emph{Proceedings of the 2016 ACM on
  international conference on multimedia retrieval}.\hskip 1em plus 0.5em minus
  0.4em\relax ACM, 2016, pp. 143--150.

\bibitem{krapac11}
\BIBentryALTinterwordspacing
J.~Krapac, J.~Verbeek, and F.~Jurie, ``\BIBforeignlanguage{English}{{Modeling
  Spatial Layout with Fisher Vectors for Image Categorization}},'' in
  \emph{\BIBforeignlanguage{English}{ICCV}}.\hskip 1em plus 0.5em minus
  0.4em\relax Barcelona, Spain: IEEE, Nov. 2011, pp. 1487--1494. [Online].
  Available: \url{http://hal.inria.fr/inria-00612277}
\BIBentrySTDinterwordspacing

\bibitem{krizhevsky12}
A.~Krizhevsky, I.~Sutskever, and G.~E. Hinton, ``Imagenet classification with
  deep convolutional neural networks,'' in \emph{NIPS}, F.~Pereira, C.~Burges,
  L.~Bottou, and K.~Weinberger, Eds.\hskip 1em plus 0.5em minus 0.4em\relax
  Curran Associates, Inc., 2012, pp. 1097--1105.

\bibitem{kwitt12}
R.~Kwitt, N.~Vasconcelos, and N.~Rasiwasia, ``Scene recognition on the semantic
  manifold,'' in \emph{ECCV}, ser. ECCV'12.\hskip 1em plus 0.5em minus
  0.4em\relax Berlin, Heidelberg: Springer-Verlag, 2012, pp. 359--372.

\bibitem{lampert09}
C.~H. Lampert, H.~Nickisch, and S.~Harmeling, ``Learning to detect unseen
  object classes by between-class attribute transfer,'' in \emph{CVPR}, June
  2009, pp. 951--958.

\bibitem{lampert13}
------, ``Attribute-based classification for zero-shot visual object
  categorization,'' \emph{TPAMI}, vol.~36, no.~3, pp. 453--465, 2014.

\bibitem{lazebnik06}
S.~Lazebnik, C.~Schmid, and J.~Ponce, ``Beyond bags of features: Spatial
  pyramid matching for recognizing natural scene categories,'' in \emph{CVPR},
  vol.~2, 2006, pp. 2169 -- 2178.

\bibitem{feifei-obank14}
L.-J. Li, H.~Su, Y.~Lim, and F.-F. Li, ``Object bank: An object-level image
  representation for high-level visual recognition,'' \emph{IJCV}, vol. 107,
  no.~1, pp. 20--39, 2014.

\bibitem{li17}
\BIBentryALTinterwordspacing
W.-X. Li and N.~Vasconcelos, ``Complex activity recognition via attribute
  dynamics,'' \emph{IJCV}, vol. 122, no.~2, pp. 334--370, Apr 2017. [Online].
  Available: \url{https://doi.org/10.1007/s11263-016-0918-1}
\BIBentrySTDinterwordspacing

\bibitem{li15}
Y.~Li, L.~Liu, C.~Shen, and A.~van~den Hengel, ``Mid-level deep pattern
  mining,'' in \emph{CVPR}, June 2015.

\bibitem{feifei-obank10}
E.~P.~X. Li-Jia~Li, Hao~Su and L.~Fei-Fei, ``Object bank: A high-level image
  representation for scene classification and semantic feature
  sparsification,'' in \emph{NIPS}, Vancouver, Canada, December 2010.

\bibitem{lin15}
T.-Y. Lin, A.~RoyChowdhury, and S.~Maji, ``Bilinear cnn models for fine-grained
  visual recognition,'' in \emph{International Conference on Computer Vision
  (ICCV)}, 2015.

\bibitem{liu_attr11}
J.~Liu, B.~Kuipers, and S.~Savarese, ``Recognizing human actions by
  attributes,'' in \emph{CVPR 2011}, June 2011, pp. 3337--3344.

\bibitem{liu14}
L.~Liu, C.~Shen, L.~Wang, A.~Hengel, and C.~Wang, ``Encoding high dimensional
  local features by sparse coding based fisher vectors,'' in \emph{Advances in
  Neural Information Processing Systems 27}, 2014, pp. 1143--1151.

\bibitem{liu16}
L.~Liu, P.~Wang, C.~Shen, L.~Wang, A.~van~den Hengel, C.~Wang, and H.~T. Shen,
  ``Compositional model based fisher vector coding for image classification,''
  \emph{CoRR}, vol. abs/1601.04143, 2016.

\bibitem{minka00}
T.~P. Minka, ``Estimating a dirichlet distribution,'' Tech. Rep., 2000.

\bibitem{perronnin07}
F.~Perronnin and C.~Dance, ``Fisher kernels on visual vocabularies for image
  categorization,'' in \emph{CVPR}, 2007, pp. 1--8.

\bibitem{perronnin10}
F.~Perronnin, J.~S\'{a}nchez, and T.~Mensink, ``Improving the fisher kernel for
  large-scale image classification,'' in \emph{ECCV}, ser. ECCV'10, 2010, pp.
  143--156.

\bibitem{dset:MITIndoor}
A.~Quattoni and A.~Torralba, ``Recognizing indoor scenes,'' \emph{CVPR},
  vol.~0, pp. 413--420, 2009.

\bibitem{rabiner}
L.~R. Rabiner, ``A tutorial on hidden markov models and selected applications
  in speech recognition,'' \emph{Proceedings of the IEEE}, vol.~77, no.~2, pp.
  257--286, Feb 1989.

\bibitem{rasiwasia07}
N.~Rasiwasia, P.~Moreno, and N.~Vasconcelos, ``Bridging the gap: Query by
  semantic example,'' \emph{Multimedia, IEEE Transactions on}, vol.~9, no.~5,
  pp. 923--938, 2007.

\bibitem{rasiwasia11}
N.~Rasiwasia and N.~Vasconcelos, ``Holistic context models for visual
  recognition,'' \emph{TPAMI}, vol.~34, no.~5, pp. 902--917, May 2012.

\bibitem{rasiwasia08}
------, ``Scene classification with low-dimensional semantic spaces and weak
  supervision,'' in \emph{CVPR}, 2008.

\bibitem{ren15}
S.~Ren, K.~He, R.~Girshick, and J.~Sun, ``Faster {R-CNN}: Towards real-time
  object detection with region proposal networks,'' in \emph{NIPS}, 2015.

\bibitem{sanchez13}
J.~S{\'a}nchez, F.~Perronnin, T.~Mensink, and J.~J. Verbeek, ``Image
  classification with the fisher vector: Theory and practice,'' \emph{IJCV},
  vol. 105, no.~3, pp. 222--245, 2013.

\bibitem{simonyan14}
K.~Simonyan and A.~Zisserman, ``Very deep convolutional networks for
  large-scale image recognition,'' \emph{CoRR}, vol. abs/1409.1556, 2014.

\bibitem{simonyan2014very}
------, ``Very deep convolutional networks for large-scale image recognition,''
  \emph{arXiv preprint arXiv:1409.1556}, 2014.

\bibitem{su12}
Y.~Su and F.~Jurie, ``Improving image classification using semantic
  attributes,'' \emph{IJCV}, vol. 100, no.~1, pp. 59--77, 2012.

\bibitem{szegedy14}
\BIBentryALTinterwordspacing
C.~Szegedy, W.~Liu, Y.~Jia, P.~Sermanet, S.~Reed, D.~Anguelov, D.~Erhan,
  V.~Vanhoucke, and A.~Rabinovich, ``Going deeper with convolutions,'' in
  \emph{CVPR}, 2015. [Online]. Available: \url{http://arxiv.org/abs/1409.4842}
\BIBentrySTDinterwordspacing

\bibitem{tanaka13}
M.~Tanaka, A.~Torii, and M.~Okutomi, ``Fisher vector based on full-covariance
  gaussian mixture model,'' \emph{IPSJ Transactions on Computer Vision and
  Applications}, vol.~5, pp. 50--54, 2013.

\bibitem{torralba11}
A.~Torralba and A.~A. Efros, ``Unbiased look at dataset bias,'' in
  \emph{CVPR'11}, June 2011.

\bibitem{torresani10}
\BIBentryALTinterwordspacing
L.~Torresani, M.~Szummer, and A.~Fitzgibbon, ``Efficient object category
  recognition using classemes,'' in \emph{ECCV}, Sep. 2010, pp. 776--789.
  [Online]. Available:
  \url{\url{http://research.microsoft.com/pubs/136846/TorresaniSzummerFitzgibbon-classemes-eccv10.pdf}}
\BIBentrySTDinterwordspacing

\bibitem{nuno00}
N.~Vasconcelos and A.~Lippman, ``A probabilistic architecture for content-based
  image retrieval,'' in \emph{CVPR}, 2000, pp. 216--221.

\bibitem{verbeek06}
J.~Verbeek, ``{Learning nonlinear image manifolds by global alignment of local
  linear models},'' \emph{{TPAMI}}, vol.~28, no.~8, pp. 1236--1250, Aug. 2006.

\bibitem{vogel07}
\BIBentryALTinterwordspacing
J.~Vogel and B.~Schiele, ``Semantic modeling of natural scenes for
  content-based image retrieval,'' \emph{IJCV}, vol.~72, no.~2, pp. 133--157,
  Apr. 2007. [Online]. Available:
  \url{http://dx.doi.org/10.1007/s11263-006-8614-1}
\BIBentrySTDinterwordspacing

\bibitem{wu15}
R.~Wu, B.~Wang, W.~Wang, and Y.~Yu, ``Harvesting discriminative meta objects
  with deep cnn features for scene classification,'' in \emph{ICCV}, December
  2015.

\bibitem{dset:MITSUN}
J.~Xiao, J.~Hays, K.~Ehinger, A.~Oliva, and A.~Torralba, ``Sun database:
  Large-scale scene recognition from abbey to zoo,'' in \emph{CVPR}, 2010, pp.
  3485--3492.

\bibitem{xu2015show}
K.~Xu, J.~Ba, R.~Kiros, K.~Cho, A.~Courville, R.~Salakhudinov, R.~Zemel, and
  Y.~Bengio, ``Show, attend and tell: Neural image caption generation with
  visual attention,'' in \emph{ICML}, 2015, pp. 2048--2057.

\bibitem{yang09}
J.~Yang, K.~Yu, Y.~Gong, and T.~Huang, ``Linear spatial pyramid matching using
  sparse coding for image classification,'' in \emph{CVPR}, 2009.

\bibitem{yang15}
S.~Yang and D.~Ramanan, ``Multi-scale recognition with dag-cnns,'' in
  \emph{ICCV}, December 2015.

\bibitem{zeiler14}
\BIBentryALTinterwordspacing
M.~D. Zeiler and R.~Fergus, ``Visualizing and understanding convolutional
  networks,'' in \emph{ECCV}, 2014, pp. 818--833. [Online]. Available:
  \url{http://dx.doi.org/10.1007/978-3-319-10590-1_53}
\BIBentrySTDinterwordspacing

\bibitem{zhou14}
B.~Zhou, A.~Lapedriza, J.~Xiao, A.~Torralba, and A.~Oliva, ``{Learning Deep
  Features for Scene Recognition using Places Database.}'' \emph{NIPS}, 2014.

\end{thebibliography}
%\small
%\bibliographystyle{ieee}
%\bibliography{}

%\begin{thebibliography}{1}
%\end{thebibliography}



\begin{IEEEbiography}[{\includegraphics[width=1in,height=1.25in,clip,keepaspectratio]{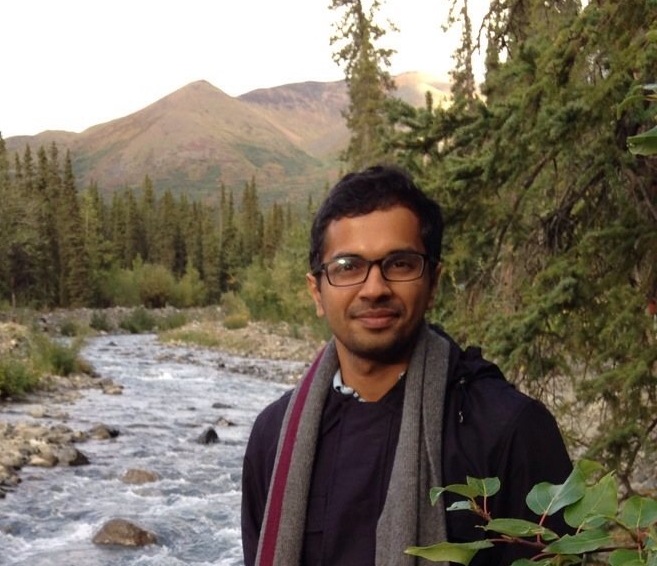}}]{Mandar Dixit}
Mandar Dixit received his PhD from the University of California at San Diego in 2018. He is currently a researcher in the computer vision technology group at Microsoft, Redmond. His research interests include in large scale visual recognition, transfer learning and meta learning.
\end{IEEEbiography}

\begin{IEEEbiography}[{\includegraphics[width=1in,height=1.25in,clip,keepaspectratio]{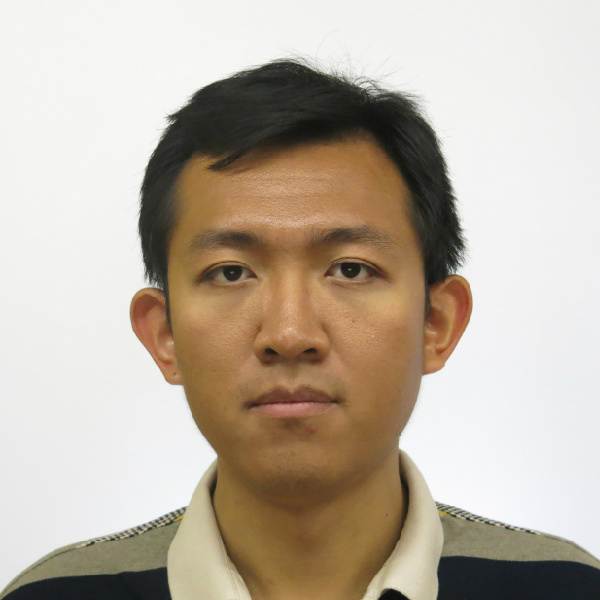}}]{Yunsheng Li}
	Yunsheng Li received his Master degreee from the University of California at San Diego in 2018. He is currently a Ph.D. student in the Statistical Visual Computing Lab of UC San Diego. His research interests include transfer learning and meta learning.
\end{IEEEbiography}

%\newpage

\begin{IEEEbiography}[{\includegraphics[width=1in,height=1.25in,clip,keepaspectratio]{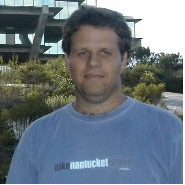}}]{Nuno Vasconcelos}
Nuno Vasconcelos received the licenciatura in electrical engineering and
computer science from the Universidade do Porto, Portugal, and the MS and PhD
degrees from the Massachusetts Institute of Technology. He is a Professor in
the Electrical and Computer Engineering Department at the University of
California, San Diego, where he heads the Statistical Visual Computing
Laboratory. He has received a NSF CAREER award, a Hellman Fellowship, 
several best paper awards, and has authored more than 150 peer-reviewed 
publications. He has been Area Chair of multiple computer vision
conferences, and is currently an Associate Editor of the IEEE
Transactions on PAMI. In 2017, he was ellected Fellow of the IEEE.\\
\end{IEEEbiography}


%\vfill


\end{document}